\documentclass[sigconf, nonacm]{acmart}

\renewcommand\footnotetextcopyrightpermission[1]{}
\usepackage{graphicx}
\usepackage{amsmath}
\usepackage{algorithm}
\usepackage{algpseudocode}
\algrenewcommand\algorithmicrequire{\textbf{Input:}}
\algrenewcommand\algorithmicensure{\textbf{Output:}}
\usepackage{booktabs}
\usepackage{placeins}
\usepackage{float}
\usepackage{multirow}
\usepackage{listings}
\usepackage{xcolor}
\usepackage{xurl}
\makeatletter
\renewcommand{\verbatim@font}{\ttfamily\footnotesize}
\makeatother
\title{TTSE: A Two-Track Online Self-Evolution \texorpdfstring{Framework\\for}{Framework for} LLM Agents}

\author{Ruimin Pei}
\email{prm21@mails.tsinghua.edu.cn}
\affiliation{\institution{Poisson Lab, Huawei}
\country{China}}

\author{Yongkang Wu}
\email{wuyongkang7@huawei.com}
\affiliation{\institution{Poisson Lab, Huawei}
\country{China}}

\author{Shangyi Zheng}
\affiliation{\institution{Poisson Lab, Huawei}
\country{China}}

\author{Yaqing Zhang}
\affiliation{\institution{Poisson Lab, Huawei}
\country{China}}

\author{Deyang Li}
\affiliation{\institution{Poisson Lab, Huawei}
\country{China}}

\author{Jianjun Tao}
\affiliation{\institution{Poisson Lab, Huawei}
\country{China}}

\author{Xinyu Zhang}
\affiliation{\institution{Poisson Lab, Huawei}
\country{China}}

\author{Xiang Zhang}
\email{zhangxiang144@huawei.com}
\affiliation{\institution{Poisson Lab, Huawei}
\country{China}}

\begin{document}
\raggedbottom

\begin{abstract}
As Large Language Model (LLM) agents are applied in continuously interactive environments, driving the evolution of their own capabilities becomes a core problem for achieving long-term autonomy. Currently, environmental knowledge is typically treated as an external fixed input rather than as part of the agent's ongoing evolution. Reinforcement learning methods usually optimize policies through environmental interaction but tend to adapt only to fixed task distributions or single environments. This paper proposes TTSE (Two-Track Self-Evolution), a dual-track online self-evolution framework that separates evolving knowledge into FACT (environmental facts, whose reliability is continuously verified through interaction evidence) and TIP (task-conditioned implementation procedures). From a decision-theoretic perspective, we decompose the agent's excess risk into environment-representation regret and conditional-execution regret, characterize the conditions under which environment-conditioned policies strictly outperform condition-agnostic policies, and bound the downstream risk in terms of FACT identification error and cross-condition mismatch cost. In practice, TTSE's ablation experiments on GDPevo validate the advantage of dual-track evolution. On the classic agent task benchmarks ALFWorld and ScienceWorld, TTSE further demonstrates superior task adaptation. Moreover, TTSE is broadly compatible with existing skill self-evolution methods; combined with the Bayesian-Agent algorithm, a single-track ablation validates the dual-track advantage, substantially improving the aggregate score across the five major domains of SOPBench over three independent repetitions. Finally, on the real end-to-end task benchmark PinchBench, TTSE is integrated into a general agent framework via retrieval-based injection and stably outperforms the baseline across three independent runs.
\end{abstract}



\maketitle
\gdef\shorttitle{TTSE: A Two-Track Online Self-Evolution Framework for LLM Agents}

\section{Introduction}

Large Language Model (LLM)-driven agents, through combining reasoning, tool use, and environmental interaction capabilities, have demonstrated powerful autonomous execution abilities in complex scenarios such as software development, web operations, scientific exploration, and embodied tasks \citep{yao2023,wang2024survey}. However, when these agents are deployed in open environments, they face challenges including continuous interaction, multi-task switching, and unknown state changes---static knowledge acquired during pretraining alone is no longer sufficient to meet the demands of long-term autonomous operation.

This challenge has spurred agent self-evolution research. However, whether it is Reflexion's reflections, Voyager's skill library, ExpeL's rules, or Bayesian-Agent's posterior updates, the core objective of existing work has consistently been to answer ``what behavior should be adopted when facing a task''---the agent's knowledge of ``what the environment is like'' is typically introduced in external static forms such as task descriptions and prompt contexts, rather than as an evolvable object that can be autonomously accumulated, verified, and revised (we survey related work in Section~\ref{sec:related}). We therefore call for a dual-track evolution mechanism.

This gap leads to practical consequences: an agent that only optimizes skills may learn experience that is effective under specific hidden conditions, producing erroneous transfer when environmental conditions change. Therefore, a long-term autonomous agent needs to simultaneously address two complementary problems: learning {how to act}, and learning {which regularities in the environment have stable validity under current interaction evidence}. Neither is sufficient alone: optimizing only skills leads to behavior that depends on outdated assumptions, while accumulating only environmental knowledge cannot be translated into execution capability.

Based on the above observations, this paper proposes {TTSE} (Two-Track Self-Evolution), a dual-track online self-evolution framework. TTSE partitions agent experiential knowledge into two independent but co-evolving tracks: {FACT} (environmental facts) and {TIP} (task-conditioned implementation procedures). The FACT track maintains ``what the environment is like'' through interpretable natural language rules---whose reliability is continuously verified through interaction evidence and updated online via count-based evidence tracking; the TIP track maintains ``how one should act''---whose effectiveness depends on environmental conditions.

This paper makes the following contributions:

\begin{enumerate}
\item We propose TTSE, a two-track online self-evolution framework whose central novelty is a structural inductive bias: environmental facts (FACT, declarative) and conditional execution policies (TIP, procedural) are maintained as two rule lifecycles that can each be independently attributed, verified, retired, and contradiction-synthesized, rather than as a single undifferentiated rule store, and we validate this bias empirically under a unified induction budget.

\item We provide a decision-theoretic characterization of dual-track knowledge: we formulate agent decision making under latent environmental conditions and exactly decompose excess risk into FACT representation regret and TIP execution regret. We further characterize when environment conditioning provides strict value and derive a dual-source risk bound connecting FACT identification errors and TIP execution quality to downstream performance; ablation experiments are conducted on GDPevo, with end-to-end evaluation on ALFWorld and ScienceWorld.

\item We demonstrate the transferability and practical compatibility of TTSE: it is compatible with existing skill self-evolution methods, and combined with the Bayesian-Agent algorithm \citep{wu2026} yields a dual-track posterior evolution framework with an aggregate improvement of +15.4pp across three repeated experiments on five major domains of SOPBench. All experiments are verified through three independent repeated runs.
\end{enumerate}

\section{Related Work}
\label{sec:related}

\subsection{Agent Self-Evolution and Skill Lifecycle Management}

Recent agent self-evolution research has progressively shifted from one-shot experiential reflection toward lifecycle management of reusable knowledge. Reflexion \citep{shinn2023} generates feedback from failed trajectories via linguistic reflection; Voyager \citep{wang2023voyager} abstracts experience into a reusable skill library; ExpeL \citep{zhao2024} pioneered extracting transferable rules from task trajectories for cross-task migration. Skill-library lifecycle management has since become central: Ratchet \citep{zhang2026} addresses ``library drift'' via outcome-driven retirement with convergence guarantees; SkillX \citep{wang2026skillx} builds a three-level transferable skill base; WebXSkill \citep{wang2026webxskill} introduces executable parameterized action programs; Bayesian-Agent \citep{wu2026} achieves posterior-guided skill evolution, a recent representative advance. However, these works focus on extracting and managing ``execution skills,'' while environmental knowledge is still assumed to be externally given.

\subsection{Reinforcement Learning and World Models}

Reinforcement learning (RL) and world models offer a parametric learning route: in continual RL, \citet{liu2025} propose an online Follow-The-Leader world model that structurally avoids catastrophic forgetting; \citet{chua2026} introduce fast/slow successor features whose progressive consolidation idea aligns with our evidence weighting. For world models, \citet{richens2025} prove that agents generalizing to multi-step goal-directed tasks must have learned a predictive environment model, with WALL-E \citep{zhou2025}, RWML \citep{yu2026}, and EMPO\textsuperscript{2} \citep{liu2026} advancing neurosymbolic, action-conditioned, and hybrid world models. These works mostly encode environmental knowledge implicitly through parametric world models, which is complementary to our route of maintaining an interpretable environmental representation via explicit natural-language rules with evidence verification.

\subsection{LLM Agent Memory Systems}

LLM agent memory systems provide a direct reference for our work. MemGPT \citep{packer2023} and Generative Agents \citep{park2023} introduce hierarchical memory architectures, but focus on individual agents' memory retrieval rather than cross-task rule evolution; MemGen \citep{zhang2025} enables agents to spontaneously evolve diverse human-like memory functions through a dynamic generative memory framework; AI-Agent School \citep{jin2025} adopts an experience-knowledge dual-memory architecture that directly supports the necessity of knowledge separation; $\tau$-Knowledge \citep{shi2026} systematically studies agent knowledge acquisition and utilization in open environments. Regarding the theoretical taxonomy of memory types, CoALA \citep{sumers2023} distinguishes semantic from procedural memory at the conceptual level; MIRIX \citep{wang2025mirix} realizes these as separate stores under coordinated management, but targets long-term personalized memory and information recall; Mem$^p$ \citep{fang2026} focuses on the online learning of procedural memory. These works either remain at the conceptual level or study a single memory type in isolation, and environmental knowledge is still treated as external input rather than a co-evolving object.

Concurrent work begins to operationalize this separation: MSCE \citep{tang2026msce} partitions agent memory into grounded traces and reusable procedural policies versus declarative environmental cognition, crystallizing evidence-backed policies into callable skills. TTSE differs by formalizing this split as an excess-risk decomposition and surfacing latent conditions from rule conflicts rather than inducing declarative knowledge by abstraction. To our knowledge, TTSE is the first work to implement environmental facts and conditional execution policies as two co-evolving online experience banks driven by task trajectories, equipped with failure attribution (Blame/Retire) and conflicting-condition synthesis (Contradict/Synthesize).

\subsection{Cognitive Science and Dual-System Memory}

The inspiration for our dual-track design comes from cognitive science. \citet{anderson1983} distinguished {declarative} (``what the world is'') from {procedural} (``how to do things'') knowledge in ACT-R, and \citet{squire1992} confirmed the biological basis of this separation---the two types of knowledge differ in brain region, learning rate, and retrieval mode. This transfer from human cognition to LLM-based agents is more than analogy: because large language models are trained on human-produced corpora, prior work reports that their internal representations share functional correspondences with human language and cognitive processing \citep{schrimpf2021}. We therefore adopt the declarative--procedural distinction as a plausible architectural prior rather than an arbitrary design choice, without claiming that the models encode this distinction as an identifiable internal structure. Yet this distinction has rarely been formally characterized in agent self-evolution; TTSE operationalizes it as two co-evolving online experience banks driven by interaction evidence, with a decision-theoretic characterization (Section~\ref{sec:theory}).

\section{System Architecture}

\subsection{Dual-Track Knowledge Representation}

TTSE organizes knowledge extracted from interaction experience into two independent tracks: {FACT} (environmental facts) and {TIP} (task-conditioned implementation procedures). The motivation for this separation stems from the observation of essential differences in the cognitive properties of these two types of knowledge.

FACT rules describe declarative properties of the environment---statements about ``what the world is like,'' such as ``objects inside closed containers are not visible until the container is opened'' or ``the same object type can have multiple instances.''

TIP rules describe conditional action strategies---procedural knowledge about ``what behavior to adopt under what conditions,'' such as ``under condition: the task requires multiple objects of the same type, after placing the first one, continue searching for remaining objects.''

\begin{figure*}[!t]
\centering
\includegraphics[width=0.85\textwidth]{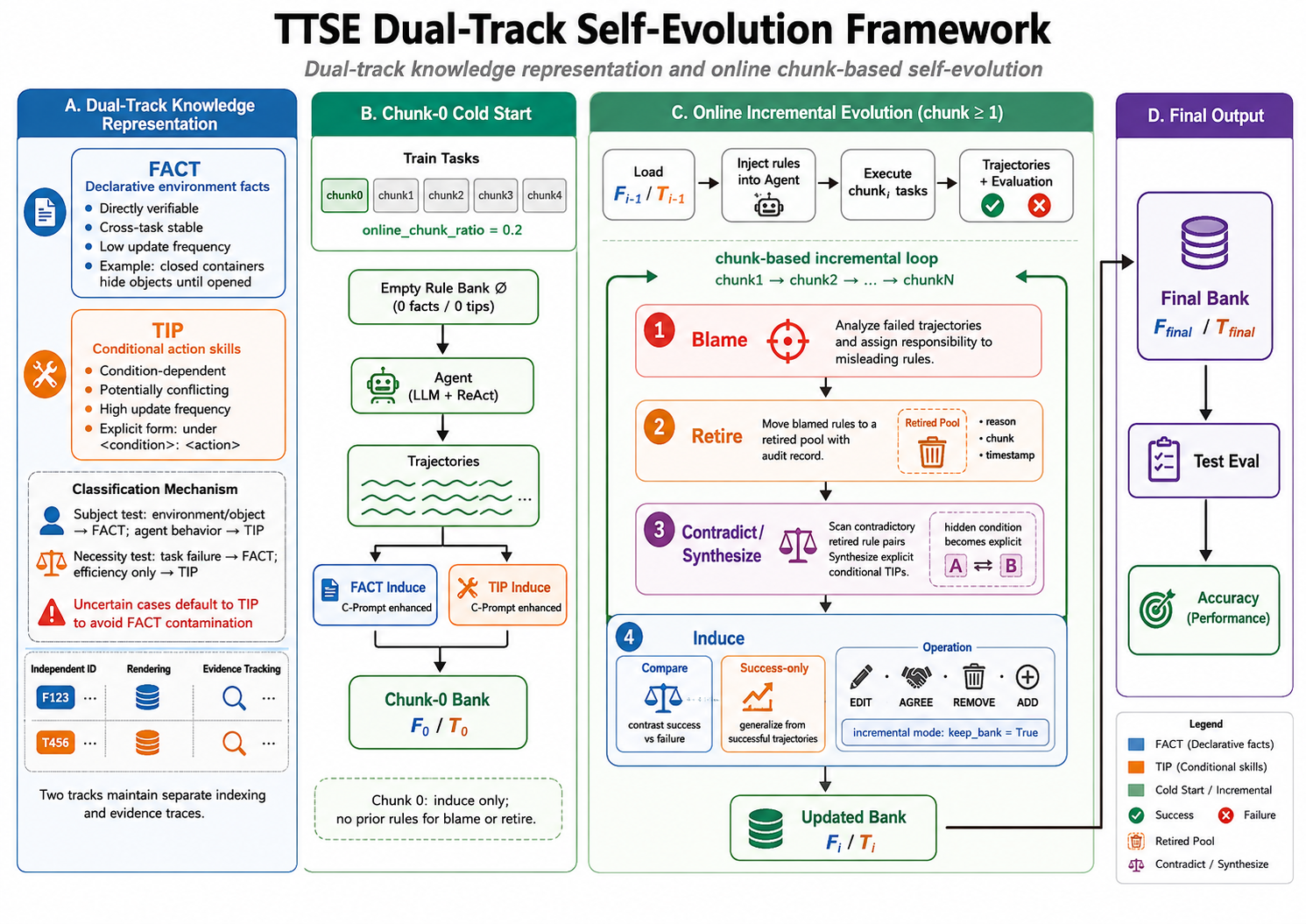}
\caption{The TTSE dual-track self-evolution framework. Knowledge is separated into FACT (environmental facts) and TIP (task-conditioned implementation procedures) channels, each with independent numbering, rendering, and evidence-tracking mechanisms. The online evolution pipeline cyclically executes Blame $\rightarrow$ Retire $\rightarrow$ Contradict/Synthesize $\rightarrow$ Induce across multiple training chunks.}
\label{fig:architecture}
\end{figure*}

\subsubsection{Classification Mechanism}

The classification of FACT versus TIP is achieved through a {dual-judgment method} in the prompt template, without requiring additional classifiers or labeled data. For each candidate rule induced by the LLM, two judgments are executed:

\begin{itemize}
\item {Subject Test}: Is the subject of the rule the environment/object ($\rightarrow$FACT) or the agent itself ($\rightarrow$TIP)?
\item {Necessity Test}: Does violating the rule lead to direct task failure ($\rightarrow$FACT, hard constraint) or only reduced efficiency ($\rightarrow$TIP, soft heuristic)?
\end{itemize}

When uncertain, the default classification is TIP. The key role of this conservative strategy is to prevent {FACT contamination}---the erroneous labeling of speculative generalizations that are only valid under specific conditions as universal truths, thereby eroding the credibility of the FACT library. For example, the rule ``washing cloth objects with a sinkbasin may fail'' describes a conditional heuristic rather than an environmental law, and must be classified as TIP.

Each TIP must follow the explicit conditional format \texttt{under <condition>: <action>}, making its dependent hidden context visible at the representation level, thereby laying the foundation for subsequent contradiction detection and condition synthesis.

\subsection{Online Evolution Pipeline}

The core of TTSE's self-evolution capability is a {chunk-based incremental loop}. Training tasks are divided into $N$ equally sized chunks, and each chunk completion triggers one round of rule library updates. Unlike the ExpeL offline pipeline (which collects all trajectories before performing one-shot induction), TTSE's online loop enables the rule library to {continuously self-correct} as training progresses---erroneous rules discovered in early chunks can be eliminated by the blame-retire mechanism in later chunks, and conflicting rules can be synthesized into conditioned knowledge via the contradict-synthesize mechanism.

Each chunk (chunk $\geq$ 1) evolution consists of four strictly ordered phases:

{Phase 1: Blame.} For all failed task execution trajectories in the current chunk, analyze one by one whether the failure cause can be attributed to any rule injected during training. If a rule directly misled the agent's decision (e.g., suggested an operation ineffective in the current environment), that rule is marked as ``culpable.''

{Phase 2: Retire.} Attributed rules are removed from the active library and transferred to the retired pool, with appended audit records (retirement reason, chunk index, timestamp). Rules in the retired pool no longer participate in subsequent task injection, but their text and retirement reasons are preserved, providing evidence for the contradiction synthesis in the next phase.

{Phase 3: Contradict/Synthesize.} When the retired pool has accumulated sufficient entries, the LLM is asked to scan the pool for contradictory rule pairs---rules that give opposite advice about the same environmental property. Such contradictions are strong signals of hidden environmental conditions: each rule's advice may be correct under the conditions from which it was induced, but the explicit characterization of applicable conditions is missing. The core operation of Contradict/Synthesize is to {explicitly surface the missing conditions}: a pair of contradictory rules is synthesized into a single TIP with explicit conditions (\texttt{under <condition>: <action>}), enabling the agent to select the correct behavioral branch according to current conditions in subsequent tasks.

{Phase 4: Induce.} Based on the complete experience (success and failure trajectories) of the current chunk, the LLM performs incremental updates to the existing rule library. Induction proceeds in two stages: (1) \emph{compare} stage---pair and contrast success and failure trajectories for the same task, distilling improvements from ``why success succeeded and why failure failed''; (2) \emph{success-only} stage---induce reusable general strategies from all successful trajectories. The incremental mode (\texttt{keep\_bank=True}) ensures that existing knowledge is not cleared; the LLM performs local optimization on the current rule library through four operations: EDIT (refine), AGREE (confirm), REMOVE (eliminate), ADD (insert).

{Cold Start (Chunk 0).} The first chunk injects no rules---the agent runs with zero knowledge and establishes the initial rule set through pure induction from the first batch of experience. This ensures that the discovery of initial rules is not influenced by existing biases.

\section{Theoretical Framework: Error Decomposition for Environment Representation and Conditional Execution}
\label{sec:theory}

\subsection{Problem Setting}

We formulate agent self-evolution as decision making under latent environmental conditions. Let $o \in \mathcal{O}$ denote the observable task context, including the task instruction and the interaction history available before a decision is made. Let $z \in \mathcal{Z}$ denote a latent environmental condition that affects which action is appropriate. For example, visually similar tasks may require different operations because of differences in object states, tool semantics, API behavior, or domain-specific constraints.

The agent selects an action $a \in \mathcal{A}$ and incurs a bounded loss
\[
\ell(o,z,a) \in [0,1].
\]
Lower loss corresponds to more successful task execution.

TTSE decomposes experiential knowledge into two functional components. The FACT channel constructs an environment representation
\[
F:\mathcal{O}\rightarrow\mathcal{S},
\]
where $F(o)$ summarizes decision-relevant properties inferred from the interaction context. The TIP channel implements a conditional policy
\[
\pi:\mathcal{O}\times\mathcal{S}\rightarrow\Delta(\mathcal{A}),
\]
where $\Delta(\mathcal{A})$ denotes the set of probability distributions over actions. Together, a FACT--TIP pair $(F,\pi)$ induces the expected risk
\begin{equation}
R(F,\pi)
=
\mathbb{E}_{(o,z)\sim\mathcal{D}}
\mathbb{E}_{a\sim\pi(\cdot\mid o,F(o))}
\left[
\ell(o,z,a)
\right].
\tag{1}
\end{equation}

This abstraction does not assume that natural-language FACTs recover the complete physical state of the environment. Instead, $F(o)$ represents the decision-relevant environmental information preserved by the FACT bank and exposed to the execution policy.

\subsection{Representation--Execution Error Decomposition}

For a fixed environment representation $F$, define the best conditional policy available in the TIP policy class $\Pi$ as
\begin{equation}
\pi_F^\star
\in
\arg\min_{\pi\in\Pi} R(F,\pi).
\tag{2}
\end{equation}

We also define an oracle policy that directly observes the true environmental condition:
\begin{equation}
R^\star
=
\inf_{\rho\in\Pi_{\mathrm{oracle}}}
\mathbb{E}_{(o,z)\sim\mathcal{D}}
\mathbb{E}_{a\sim\rho(\cdot\mid o,z)}
\left[
\ell(o,z,a)
\right].
\tag{3}
\end{equation}
The oracle class is assumed to contain policies that can ignore $z$ and use only $F(o)$; therefore, access to the true condition cannot increase the minimum achievable risk.

\paragraph{Proposition 1 (Representation--execution decomposition).}
For any FACT representation $F$ and TIP policy $\pi$, the excess risk relative to the environment-aware oracle can be decomposed exactly as
\begin{equation}
\begin{aligned}
R(F,\pi)-R^\star
={}&
\underbrace{
R(F,\pi)-R(F,\pi_F^\star)
}_{\mathcal{E}_{\mathrm{TIP}}(F,\pi)}
\\
&+
\underbrace{
R(F,\pi_F^\star)-R^\star
}_{\mathcal{E}_{\mathrm{FACT}}(F)}.
\end{aligned}
\tag{4}
\end{equation}

Here, $\mathcal{E}_{\mathrm{TIP}}$ is the execution regret remaining under the current environment representation, whereas $\mathcal{E}_{\mathrm{FACT}}$ is the representation regret that cannot be eliminated by improving execution alone. Both terms are non-negative.

Proposition~1 identifies two conceptually distinct sources of failure. Even with a sufficiently informative FACT representation, the agent may fail because it cannot reliably select or execute the appropriate conditional action. Conversely, even an effective TIP policy cannot fully compensate for an environment representation that omits decision-relevant distinctions.

\subsection{Value of Environment-Conditioned Policies}

To characterize when environmental knowledge provides strict value, consider two policy classes. An environment-agnostic policy uses only the observable task context:
\[
\Pi_{\mathrm{agn}}
=
\left\{
\rho:\mathcal{O}\rightarrow\Delta(\mathcal{A})
\right\},
\]
whereas an environment-conditioned policy may additionally depend on the true condition:
\[
\Pi_{\mathrm{cond}}
=
\left\{
\rho:\mathcal{O}\times\mathcal{Z}
\rightarrow\Delta(\mathcal{A})
\right\}.
\]

Their optimal risks are
\begin{equation}
R_{\mathrm{agn}}^\star
=
\inf_{\rho\in\Pi_{\mathrm{agn}}}
\mathbb{E}\left[
\ell(o,z,a)
\right],
\tag{5}
\end{equation}
and
\begin{equation}
R_{\mathrm{cond}}^\star
=
\inf_{\rho\in\Pi_{\mathrm{cond}}}
\mathbb{E}\left[
\ell(o,z,a)
\right].
\tag{6}
\end{equation}

Define the value of environmental conditioning as
\begin{equation}
H
=
R_{\mathrm{agn}}^\star
-
R_{\mathrm{cond}}^\star.
\tag{7}
\end{equation}

\paragraph{Proposition 2 (Value of environmental conditioning).}
For any task distribution and bounded loss,
\begin{equation}
H\geq 0.
\tag{8}
\end{equation}
Moreover, $H>0$ whenever, with positive probability, the same observable context is compatible with multiple environmental conditions that require different optimal actions with a positive loss margin.

The result follows because every environment-agnostic policy is a special case of an environment-conditioned policy. The inequality is strict when no single action distribution can be optimal across all conditions consistent with the same observable context.

Proposition~2 clarifies that FACT is not expected to provide equal value in every task domain. When environments are effectively homogeneous, or when one action is optimal under all conditions, environmental conditioning provides little additional benefit. Its value increases when superficially similar situations require conflicting actions.

\paragraph{Corollary 1 (Binary conflict environment).}
Suppose two environmental conditions are equally likely and require mutually incompatible unique actions, while the observable context alone does not distinguish the conditions. Under zero--one loss, any environment-agnostic policy has success probability at most $1/2$. A system that correctly identifies the condition and executes the corresponding conditional action can achieve success probability $1$.

Contradict/Synthesize implements this form of conditioning at the representation level: conflicting unconditional rules are converted into an explicit condition--action TIP rather than being treated as a single universally valid instruction.

\subsection{A Dual-Source Excess-Risk Bound}

We next connect FACT identification quality to downstream decision risk. Assume that $\mathcal{Z}$ is finite and that the FACT representation outputs a predicted condition
\[
\widehat{z}=F(o)\in\mathcal{Z}.
\]
Let
\begin{equation}
a^\star(o,z)
\in
\arg\min_{a\in\mathcal{A}}\ell(o,z,a)
\tag{9}
\end{equation}
denote an oracle action for condition $z$. Define the FACT identification error
\begin{equation}
\eta_F
=
\Pr\left[F(o)\neq z\right],
\tag{10}
\end{equation}
and the maximum cost of applying an action optimized for an incorrect condition as
\begin{equation}
\Delta_{\max}
=
\sup_{o,z,z'}
\left[
\ell\bigl(o,z,a^\star(o,z')\bigr)
-
\ell\bigl(o,z,a^\star(o,z)\bigr)
\right]_{+}.
\tag{11}
\end{equation}
Because $\ell\in[0,1]$, we have $0\leq\Delta_{\max}\leq 1$.

\paragraph{Proposition 3 (Dual-source excess-risk bound).}
Assume that the TIP policy class can realize the plug-in conditional policy $a^\star(o,F(o))$. Then
\begin{equation}
\mathcal{E}_{\mathrm{FACT}}(F)
\leq
\eta_F\Delta_{\max},
\tag{12}
\end{equation}
and consequently
\begin{equation}
R(F,\pi)-R^\star
\leq
\mathcal{E}_{\mathrm{TIP}}(F,\pi)
+
\eta_F\Delta_{\max}.
\tag{13}
\end{equation}

This bound gives two complementary routes to performance improvement. TIP evolution reduces execution regret under the current representation. FACT evolution reduces the probability of selecting a behavioral branch corresponding to the wrong environmental condition. The impact of a FACT error is amplified when cross-condition action mismatch is costly, as quantified by $\Delta_{\max}$.

\subsection{Implications for TTSE}

The theoretical quantities above should be interpreted as functional targets rather than as directly observed optimization objectives. FACT induction and evidence tracking aim to preserve more decision-relevant environmental distinctions, thereby reducing representation regret. Conditional TIP induction and rule rendering aim to reduce execution regret under the inferred context. Blame and Retire provide targeted credit assignment for misleading rules, while Contradict/Synthesize converts conflicting unconditional advice into explicitly conditioned behavior.

Accordingly, TTSE's algorithmic mechanisms can be viewed as practical approximations for separately managing the two error terms in Eq.~(4). The theory does not require either channel to improve monotonically after every interaction, nor does it claim that the implemented update procedure necessarily converges to the oracle risk.

\paragraph{Scope of the analysis.}
Our results do not claim that every unified memory system is less expressive than a dual-track system. A sufficiently expressive single-bank representation could, in principle, encode both environmental properties and conditional policies. The theoretical role of TTSE is instead to make two sources of decision error explicit and separately manageable. FACT/TIP separation therefore serves as a modular inductive bias for knowledge induction, rendering, verification, and credit assignment, rather than as a universal expressivity advantage. Full proofs are provided in Appendix~A.

\section{Experiments}
\label{sec:exp}

This paper evaluates TTSE through four complementary research questions, with the corresponding experiments developed across Section~\ref{sec:exp} (Table~\ref{tab:logic}). First, \textbf{GDPevo} examines dual-track complementarity via FACT-only, TIP-only, and FACT+TIP ablations (Mechanism). Second, \textbf{SOPBench} compares unified single-track and dual-track representations under the same Bayesian lifecycle management, examining whether the dual-track advantage transfers to an external skill-management mechanism (Transferability). Third, \textbf{ALFWorld} and \textbf{ScienceWorld} evaluate the end-to-end task-adaptation capability of the complete TTSE system (End-to-end effectiveness). Fourth, \textbf{PinchBench} does not re-isolate the marginal contribution of FACT/TIP classification; instead, it examines whether TTSE can be integrated non-invasively into a real agent runtime and retain its performance gains in open, multi-step tool-calling environments (Deployment).

\begin{table}[!t]
\centering
\caption{The four complementary research questions and the corresponding benchmarks.}
\label{tab:logic}
\setlength{\tabcolsep}{3pt}
\begin{tabular}{p{0.20\linewidth}p{0.30\linewidth}p{0.36\linewidth}}
\toprule
Evidence level & Benchmark & Research question \\
\midrule
Mechanism & GDPevo & Dual- vs.\ single-track ablation \\
Transferability & SOPBench & Ablation and compatibility with other mechanisms \\
End-to-end & ALFWorld / ScienceWorld & Task adaptation on classic benchmarks \\
Deployment & PinchBench & Harness compatibility and practical utility \\
\bottomrule
\end{tabular}
\end{table}

Section~\ref{sec:threebm} presents ALFWorld, ScienceWorld, and GDPevo; Sections~\ref{sec:sop} and~\ref{sec:pinch} report the SOPBench fusion and the PinchBench deployment experiments.

\subsection{Three Classic Benchmarks: ALFWorld, ScienceWorld, and GDPevo}
\label{sec:threebm}

\subsubsection{ALFWorld}
\label{sec:alf}

ALFWorld \citep{alfworld} is an embodied agent benchmark containing 134 valid unseen tasks spanning 6 types. The agent must use natural language instructions to navigate rooms, manipulate objects, and apply state changes (cleaning, heating, cooling). We randomly sample 500 tasks from the full training set as the training set, with 5 chunks (chunk\_ratio=0.2), 50 shard parallelism, and a maximum of 50 steps per task. Evaluation uses the rule library from the final chunk. Two LLMs each run independently three times: DeepSeek-V3.2 and GLM-5 (both in thinking mode). The step count metric is pure action steps (excluding think steps); three-run means are averaged. This setup matches the baselines' evaluation protocol---same backbones, same unseen-task evaluation split, same 50-step budget, and no privileged information---and the rule library used for evaluation is the one retained after the final training chunk, so the comparison reflects deployment-realistic experience accumulation rather than a per-task oracle.

{Comparison with Existing Methods:}

\begin{table}[!t]
\centering
\caption{ALFWorld success rate (SR) and step count comparison. All values are three-run means. GraSP and the ReAct/ExpeL results reproduced therein are from the original GraSP paper \citep{xia2026}; SkillOpt \citep{ju2026} and SkillX \citep{wang2026skillx} are our reproductions under the same setting. TTSE achieves the best result on both backbones, significantly outperforming skill-optimization baselines such as SkillOpt and SkillX.}
\label{tab:alfworld}
\setlength{\tabcolsep}{3pt}
\begin{tabular}{lcccc}
\toprule
\multirow{2}{*}{\bf Method} & \multicolumn{2}{c}{\bf DS-V3.2} & \multicolumn{2}{c}{\bf GLM-5} \\
\cmidrule(lr){2-3}\cmidrule(lr){4-5}
 & SR$\uparrow$ & Steps$\downarrow$ & SR$\uparrow$ & Steps$\downarrow$ \\
\midrule
ReAct \citep{yao2023} & 69.4 & 19.3 & 64.6 & 14.6 \\
ExpeL \citep{zhao2024} & 76.1 & 17.4 & 72.6 & 11.4 \\
GraSP \citep{xia2026} & 83.6 & 14.8 & 94.9 & 10.3 \\
SkillOpt \citep{ju2026} & 68.7 & 13.3 & 62.7 & 15.1 \\
SkillX \citep{wang2026skillx} & 56.5 & 11.2 & 88.3 & 9.2 \\
{TTSE (Ours)} & {\bf 97.8} & {\bf 10.0} & {\bf 98.51} & {\bf 9.2} \\
\bottomrule
\end{tabular}
\end{table}

Under the same backbone and evaluation split, TTSE outperforms previously reported results: 97.8\% on DeepSeek-V3.2 vs.\ GraSP's 83.6\% (+14.2pp), and 98.51\% on GLM-5 vs.\ GraSP's 94.9\% (+3.6pp). We attribute this efficiency advantage to the dual-channel rendering mechanism: FACT rules provide environmental priors that reduce exploratory trial-and-error, while TIP rules provide verified operational procedures that reduce ineffective actions.

\subsubsection{ScienceWorld}

ScienceWorld \citep{scienceworld} is a textual science experiment simulation containing 30 task types. Tasks span chemistry experiments, biological classification, physical measurement, and genetics problems. The agent must navigate simulated science laboratories, use instruments, and conduct experiments. We train online using GLM-5 (thinking): 30 shards, max 50 steps, 5 chunks. After training, three independent evaluations are conducted.

{Three-Run Results:}

\begin{table}[!t]
\centering
\caption{ScienceWorld six-run comparison. TTSE achieves a substantial improvement over the corresponding baseline.}
\label{tab:scienceworld}
\setlength{\tabcolsep}{1.2mm}
\begin{tabular}{lcc}
\toprule
& {MACRO Reward} & {Native Success} \\
\midrule
Baseline R1 & 0.7335 & 55.47\% \\
Baseline R2 & 0.7100 & 52.45\% \\
Baseline R3 & 0.7177 & 51.32\% \\
{Baseline Mean $\pm\sigma$} & {0.7204$\pm$0.010} & {53.08\%$\pm$1.74} \\
\midrule
TTSE R0 & 0.8634 & 77.36\% \\
TTSE R1 & 0.8366 & 69.43\% \\
TTSE R2 & 0.8602 & 73.58\% \\
{TTSE Mean $\pm\sigma$} & {0.8534$\pm$0.012} & {73.46\%$\pm$3.2} \\
\midrule
{$\Delta$ vs Baseline} & {+0.1330} & {+20.38pp} \\
\bottomrule
\end{tabular}
\end{table}

\subsubsection{GDPevo Ablation}
\label{sec:gdp}

GDPevo \citep{gdpevo} is a multi-domain enterprise operations benchmark covering 12 task groups across CRM, ERP, finance, and HR domains. We conduct {4-condition ablation experiments} on GDPevo using GLM-5 (thinking mode), with each condition independently repeated three times. The baseline is the classic ReAct framework; the experimental groups differ only in the dual-track prompt wording. The \texttt{full} group explicitly instructs the extraction of both tracks into separate banks; the \texttt{tips} and \texttt{facts} groups explicitly require extraction of only tips or facts in the induction prompt.

\begin{table}[!t]
\centering
\footnotesize
\caption{GDPevo four-condition ablation (three runs). All conditions are trained online except \texttt{base} (zero-shot); each condition's bank composition follows its name. \texttt{full} (FACT+TIP) mean 0.2616, 2.8$\times$ over baseline. Dual-track beats single-track tips (+2.4pp) and facts (+7.0pp).}
\label{tab:gdpevo}
\setlength{\tabcolsep}{4pt}
\begin{tabular}{lcccc}
\toprule
{Condition} & {Mean} & {R1} & {R2} & {R3} \\
\midrule
{full} (FACT+TIP) & {\bf 0.2616} & 0.2849 & 0.2528 & 0.2471 \\
tips (TIP only) & 0.2380 & 0.2312 & 0.2513 & 0.2314 \\
facts (FACT only) & 0.1918 & 0.1546 & 0.2045 & 0.2163 \\
base (zero-shot) & 0.0947 & 0.0765 & 0.0840 & 0.1235 \\
\bottomrule
\end{tabular}
\end{table}

The ablation results provide evidence for the dual-track design. \texttt{full} (FACT+TIP) achieves a three-run mean of 0.2616, an improvement of 0.1669 (approximately 2.8$\times$) over the zero-shot baseline of 0.0947. Single-track variants underperform dual-track: TIP-only achieves 0.2380 ($-2.4$pp vs.\ full) and FACT-only only 0.1918 ($-7.0$pp vs.\ full). We note that the gap between \texttt{full} and TIP-only is not accompanied by a formal significance test; the difference is consistent in direction across runs but its magnitude should be read as ablation evidence rather than a statistically certified margin. These results are consistent with Proposition~1 rather than constituting a direct verification of its latent quantities. FACT-only leaves substantial execution regret because environmental descriptions alone do not specify a reliable conditional procedure. TIP-only performs better but lacks an independently maintained representation for distinguishing environment-dependent behavioral branches. The full system can address both sources of error.

GDPevo's enterprise operation domain provides complementary validation to ALFWorld and ScienceWorld: the former tests TTSE's dual-track mechanism under controlled ablation, while the latter two test the complete system's end-to-end task adaptation. Together, the experiments show that TTSE outperforms previously reported results on ALFWorld and substantially improves over the corresponding baseline on ScienceWorld.

\subsection{SOPBench Fusion Experiment}
\label{sec:sop}

To demonstrate the generality of the dual-track paradigm, we integrate TTSE with the {Bayesian-Agent} posterior-driven skill management system \citep{wu2026}. In this fusion, Bayesian posteriors $\text{Beta}(\alpha, \beta)$ replace TTSE's heuristic +1/$-$1 counting for rule lifecycle decisions (Retire/Stable/Exploring), while TTSE's FACT/TIP representation and LLM-driven induction provide the knowledge acquisition mechanism. The knowledge base starts completely empty---all rules are induced from scratch by DeepSeek v4-flash from failure trajectories, with no hand-written seed rules. At each step, the top-8 FACT and top-8 TIP rules by posterior success probability are rendered into the agent context.

We evaluate on {SOPBench} \citep{sopbench}, selecting 5 representative domains with large sample sizes (892 tasks $\times$ 3 repetitions = 2,676 total evaluations). To isolate the contribution of FACT/TIP classification, we conduct a {single-track ablation}: all rules are induced into a single unified bank without FACT/TIP type distinction, and the induction prompt imposes no special requirements on rule categories---the LLM freely induces rules in a unified ``rule'' format, with only Bayesian posterior updates governing rule quality.

\begin{table}[!t]
\centering
\footnotesize
\caption{SOPBench ablation (three-run means). Bayesian posteriors improve accuracy; FACT/TIP classification adds +3.0pp overall.}
\label{tab:sopbench}
\setlength{\tabcolsep}{1.0mm}
\begin{tabular}{lrcccc}
\toprule
{Domain} & {Tasks} & {Base} & {Single} & {Dual} & {$\Delta$} \\
\midrule
aircraft\_insp. & 112 & 95.8\% & 97.6\% & {98.8\%} & +1.2pp \\
customer\_svc. & 156 & 97.2\% & 97.9\% & {98.9\%} & +1.0pp \\
dangerous\_goods & 274 & 77.1\% & 94.6\% & {94.0\%} & {$-$0.6pp} \\
referral\_abuse & 200 & 97.8\% & 97.7\% & {99.5\%} & +1.8pp \\
warehouse\_insp. & 150 & 26.7\% & 66.9\% & {81.3\%} & {+14.4pp} \\
\midrule
{Overall (micro)} & 892 & 79.15\% & 91.59\% & {94.58\%} & {+2.99pp} \\
\bottomrule
\end{tabular}
\end{table}

\begin{figure}[!t]
\centering
\includegraphics[width=\columnwidth]{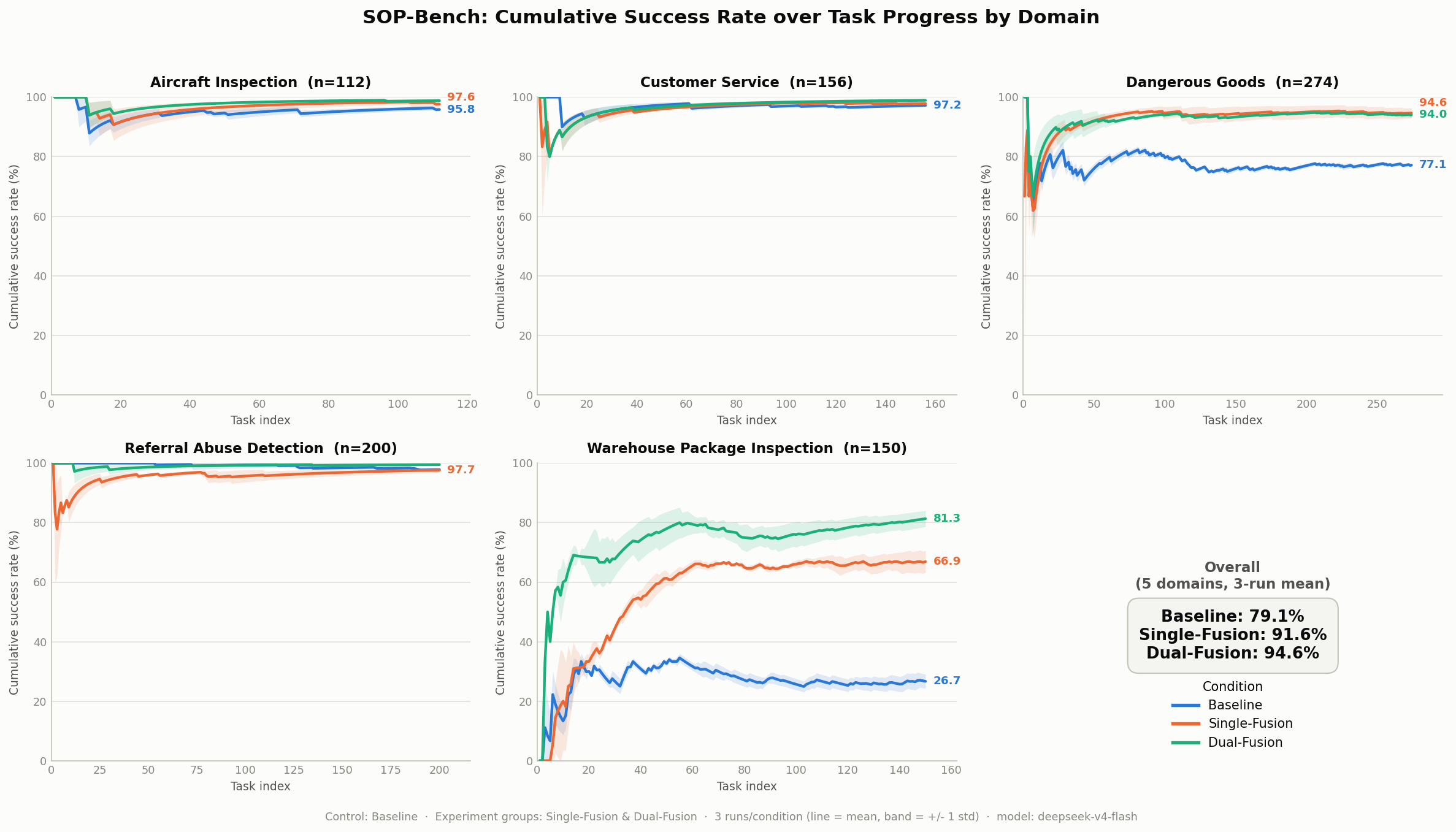}
\caption{SOPBench success rate by domain (three-run means). Dual-track (FACT+TIP) outperforms the single-track ablation overall and in four of five domains, with the largest gain in warehouse\_package\_inspection.}
\label{fig:sopbench_domain}
\end{figure}

Both single-track and dual-track systems achieve substantial improvements over the baseline (Figure~\ref{fig:sopbench_domain}). The dual-track FACT/TIP classification contributes an additional +3.0pp overall. Moreover, beyond this accuracy advantage, a deeper analysis of the induced rules reveals more fundamental structural differences (Table~\ref{tab:rules_quality}).

\begin{table}[!t]
\centering
\footnotesize
\caption{Rules quality: Single vs.\ Dual. Dual-track rules are 18\% shorter (137 vs.\ 167 chars/rule), saving context tokens.}
\label{tab:rules_quality}
\setlength{\tabcolsep}{1.2mm}
\begin{tabular}{lrrcr}
\toprule
{Domain} & \multicolumn{2}{c}{Single-track} & \multicolumn{2}{c}{Dual-track} \\
\cmidrule(lr){2-3}\cmidrule(lr){4-5}
& {Rules} & {Chars/rule} & {Rules} & {Chars/rule} \\
\midrule
aircraft\_insp. & 11 & 191 & 5 & 140 \\
customer\_svc. & 10 & 167 & 9 & 145 \\
dangerous\_goods & {52} & 158 & {59} & 128 \\
referral\_abuse & 15 & 155 & 4 & 139 \\
warehouse\_insp. & 37 & 166 & 69 & 134 \\
\midrule
{Average} & {25} & {167} & {29} & {137} \\
\bottomrule
\end{tabular}
\end{table}

Three structural findings emerge from Table~\ref{tab:rules_quality}. (1) Dual-track rules are 18\% shorter (137 vs.\ 167 chars/rule)---each injection brings immediate token savings. (2) warehouse\_package\_inspection is the most compelling case: with a very low baseline, our dual-track achieves an upward leap in improvement. (3) Shorter rules together yield approximately a 17\% reduction in rule context tokens per task injection. Overall, the dual-track representation is effective in aggregate, with its advantage most pronounced in domains where environmental rules are complex and FACT/TIP complementarity is strong.

Statistical testing further supports the reliability of the overall dual-track advantage: paired bootstrap across 892 tasks (10,000 resamples) shows that the +2.99pp overall gain is statistically significant (permutation test $p < 10^{-4}$, 95\% CI [+2.34pp, +3.65pp]), with warehouse\_package\_inspection contributing 78\% of the total gain (+14.44pp). Even excluding warehouse, dual-track retains a +0.67pp positive advantage ($p = 0.011$).

\subsection{PinchBench Deployment}
\label{sec:pinch}

It is important to clarify that the PinchBench experiment is \emph{not} designed to re-isolate the marginal contribution of FACT/TIP classification. That question is already answered by the GDPevo single-track ablation (Section~\ref{sec:threebm}) and the SOPBench unified-single-track control (Section~\ref{sec:sop}). Instead, this experiment asks a different question: whether TTSE can be integrated \emph{non-invasively into a real agent harness} and retain its end-to-end benefit in open, multi-step tool-calling task streams. The first four benchmarks (ALFWorld, ScienceWorld, GDPevo, SOPBench) all evaluate TTSE in relatively controlled task spaces; to address the deployment question, we use PinchBench as a fifth experiment. Unlike controlled sandbox tasks, PinchBench contains 147 real end-to-end tasks spanning writing, coding, research, CSV analysis, meeting minutes, log analysis, and browser automation; the agent must complete multi-step tool calls and file production, with an independent judge scoring output quality on a continuous 0--1 scale.

We integrate the TTSE dual-track module on top of a general-purpose agent framework (OpenClaw). Unlike the prompt-context whole-bank injection used earlier for fixed-category task sets, here we adopt \emph{semantic retrieval injection}: all bank experience is encoded with all-MiniLM-L6-v2 (384-dim), and each new task performs a cosine-similarity top-$K$ retrieval over its prompt (top-10 FACT and top-10 TIP), injecting only the experience most relevant to that task so that irrelevant entries do not dilute attention. FACTs are written into \texttt{ENVIRONMENT.md} (anchoring ``what the target looks like'') and TIPs into \texttt{TIPS.md} (providing ``how to do it'').

We use the bare agent (\texttt{TTSE\_ENABLE=0}, no experience injected) as the baseline and run both configurations independently three times:

\begin{table}[!t]
\centering
\footnotesize
\caption{PinchBench three independent runs (means over all 147 task scores). TTSE beats the baseline in every single run (+3.83pp on average); the run-to-run range drops from 3.33pp to 1.61pp, i.e., experience accumulation makes the agent's performance more stable and less reliant on single-run luck.}
\label{tab:pinchbench}
\setlength{\tabcolsep}{2.5pt}
\begin{tabular}{lccccc}
\toprule
Config & r0 & r1 & r2 & mean & range \\
\midrule
baseline & 0.7259 & 0.7592 & 0.7466 & 0.7439 & 3.33pp \\
TTSE & 0.7718 & 0.7879 & 0.7868 & \bf 0.7822 & \bf 1.61pp \\
per-run $\Delta$ & +4.59pp & +2.87pp & +4.02pp & \bf +3.83pp & halved \\
\bottomrule
\end{tabular}
\end{table}

It is worth emphasizing that TTSE outperforms the baseline in \emph{each} of the three independent runs (+4.59pp, +2.87pp, +4.02pp), a consistent and stable direction of gain. In one run, the number of strictly perfect (1.0) tasks rose from 65/147 to 70/147---the gain is reflected not only in the continuous-score mean but also in pushing ``almost-done'' tasks across the perfection line.

The TIP track does not merely guide basic tools: it fully interfaces with the framework's AgentSkill system. Each skill is a self-describing unit (with a \texttt{SKILL.md} manifest) that the agent lazy-loads via the \texttt{read} tool only when the skill description matches the current task. The bank's TIPs already include entries that explicitly guide the agent on when and how to invoke a particular skill (e.g., ``when a task requires installing a capability from a registry: use the \texttt{clawhub} skill''; ``when a task requires GitHub CLI operations: use the \texttt{github} skill''). This shows that TTSE follows the industry-standard skill protocol and that the TIP track already drives the agent to discover and invoke skills, interfacing seamlessly with the existing skill ecosystem.

\FloatBarrier
\section{Conclusion}

This paper presents {TTSE (Two-Track Self-Evolution)}, an online dual-track self-evolution framework for LLM agents. Its core insight is that interaction-derived knowledge naturally falls into two categories---declarative environmental facts (FACT) and task-conditioned procedures (TIP)---and conflating them causes knowledge conflicts and spurious rules. Through independent FACT and TIP channels, each with dedicated induction, rendering, deduplication, and evidence tracking, TTSE lets environmental knowledge and execution skills co-evolve. Across five benchmarks, TTSE obtains stable positive results on the overall or primary evaluation metrics, though the magnitude of the dual-track benefit varies by domain and is more pronounced in tasks with higher environmental-condition heterogeneity.

The theoretical analysis does not treat FACT/TIP separation as a universal expressivity advantage over all possible unified memories. Instead, it shows that environment representation and conditional execution correspond to distinct sources of decision error. FACT/TIP separation should therefore be understood as a modular inductive bias---not as a claim of universal expressive power---and TTSE makes these two error sources separately manageable through dedicated induction, rendering, evidence tracking, and credit-assignment mechanisms.

More fundamentally, the experiments provide broad support for dual-track FACT/TIP separation as a transferable, modular inductive bias: it works across different evidence-tracking mechanisms (heuristic counting and Bayesian posteriors), different LLM backbones (GLM-5, DeepSeek-V3.2, DeepSeek v4-flash), and diverse task domains (enterprise API auditing, embodied interaction, scientific reasoning, SOP execution, open-ended tool-calling). The GDPevo ablation provides ablation evidence for dual-track over single-track variants (the Full-vs-TIP-only gap is not formally significance-tested); ALFWorld exceeds previously reported results under the same backbone; ScienceWorld improves substantially over its baseline; the SOPBench fusion confirms compatibility with posterior-driven lifecycle management; and PinchBench shows non-invasive integration into a real agent harness with retained end-to-end gains. Taken together, the declarative/procedural separation of knowledge is a broadly effective design principle across the evaluated settings. This has direct design implications: mainstream agent frameworks typically maintain only a procedural knowledge base (\texttt{skill.md}), lacking an independent, co-evolving environmental-knowledge representation (\texttt{environment.md}). Treating environmental facts as first-class knowledge that co-evolves with skills is a key architectural decision for long-term agent autonomy.

\section{Limitations and Future Work}

Although TTSE demonstrates robust gains across multiple benchmarks, this work has several limitations. First, TTSE's rule bank starts from a blank cold start, so the agent's behavior in the initial chunk relies entirely on the LLM's zero-shot ability; in domains where the LLM itself is insufficient, a promising direction is to introduce a small number of high-quality human examples as seed rules or to use a stronger LLM for initial bootstrapping. Second, rule induction in this paper relies entirely on the LLM, and the LLM may produce spurious or overly specialized rules on low-quality trajectories; the current system filters these post-hoc via evidence counting and the Subject/Necessity Test dual classification, but the induction stage itself lacks a structured verification mechanism---coupling the induction process with execution verification (e.g., requiring a new rule to pass on held-out verification tasks before entering the bank) may further improve rule quality, though as base-model capabilities continue to improve this issue is expected to be naturally mitigated, and non-LLM rule-distillation mechanisms can also interface seamlessly with our dual-track mechanism.

\section{Ethical Considerations}

This work does not involve human subjects or personal data. All experiments use publicly available benchmarks (ALFWorld, ScienceWorld, GDPevo, SOPBench, and PinchBench) with synthetic or non-sensitive agent trajectories; no real user data is collected, stored, or released.

A beneficial side effect of TTSE is that it makes an agent's accumulated knowledge more explicit and auditable. Both tracks are maintained as human-readable natural-language rules whose reliability is tracked through interaction-derived evidence counts, so the basis for any induced behavior can be inspected and revoked (retired) rather than being opaque in latent parameters; this transparency is a deliberate design choice that facilitates oversight. As with any method that improves autonomous agent capability, dual-use risk exists---more capable agents could in principle be redirected toward harmful tasks---and we encourage deploying TTSE-based agents under the same access controls, action logging, and human-in-the-loop review that govern autonomous agents generally.

We report all results as means over three independent repetitions to avoid cherry-picking favorable runs, and we disclose domain-level cases in which dual-track underperforms the single-track ablation (e.g., dangerous\_goods) rather than reporting only aggregate gains. The repeated-rollout evaluation protocol incurs additional LLM inference cost; we bound this through fixed trajectory budgets and, where applicable, batched induction that amortizes one reflection call over multiple tasks.

\bibliographystyle{ACM-Reference-Format}
\bibliography{references}

\appendix
\setcounter{secnumdepth}{3}

\section{Complete Proofs for the Representation--Execution Framework}

This appendix provides complete proofs for the theoretical results presented in Section~4. The analysis formalizes the distinction between environment representation and conditional execution. It is not intended as a convergence proof for the particular LLM-based update operators used by TTSE.

\subsection{Notation and Basic Definitions}

Let $(o,z)\sim\mathcal{D}$, where $o\in\mathcal{O}$ is the observable task context and $z\in\mathcal{Z}$ is a latent environmental condition. Let $\mathcal{A}$ be the action space and
\[
\ell:\mathcal{O}\times\mathcal{Z}\times\mathcal{A}\rightarrow[0,1]
\]
be a bounded loss function.

A FACT representation is a mapping
\[
F:\mathcal{O}\rightarrow\mathcal{S}.
\]
A TIP policy is a conditional action distribution
\[
\pi(\cdot\mid o,F(o))\in\Delta(\mathcal{A}).
\]
Their expected risk is
\begin{equation}
R(F,\pi)
=
\mathbb{E}_{(o,z)\sim\mathcal{D}}
\mathbb{E}_{a\sim\pi(\cdot\mid o,F(o))}
\left[
\ell(o,z,a)
\right].
\tag{14}
\end{equation}

Let $\Pi$ denote the admissible TIP policy class. For a fixed representation $F$, define
\begin{equation}
\pi_F^\star
\in
\arg\min_{\pi\in\Pi}R(F,\pi).
\tag{15}
\end{equation}

Let $\Pi_{\mathrm{oracle}}$ denote a policy class whose policies can condition directly on $(o,z)$. We assume that it contains all policies obtainable from $\Pi$ and $F$, because an oracle policy may ignore $z$ and output
\[
\rho(\cdot\mid o,z)=\pi(\cdot\mid o,F(o)).
\]
The oracle risk is
\begin{equation}
R^\star
=
\inf_{\rho\in\Pi_{\mathrm{oracle}}}
\mathbb{E}_{(o,z)}
\mathbb{E}_{a\sim\rho(\cdot\mid o,z)}
\left[
\ell(o,z,a)
\right].
\tag{16}
\end{equation}

\subsection{Proof of Proposition 1}

\paragraph{Proposition 1.}
For any FACT representation $F$ and TIP policy $\pi$,
\begin{equation}
\begin{aligned}
R(F,\pi)-R^\star
={}&
\mathcal{E}_{\mathrm{TIP}}(F,\pi)
+
\mathcal{E}_{\mathrm{FACT}}(F),
\end{aligned}
\tag{17}
\end{equation}
where
\begin{equation}
\mathcal{E}_{\mathrm{TIP}}(F,\pi)
=
R(F,\pi)-R(F,\pi_F^\star)
\tag{18}
\end{equation}
and
\begin{equation}
\mathcal{E}_{\mathrm{FACT}}(F)
=
R(F,\pi_F^\star)-R^\star.
\tag{19}
\end{equation}
Both terms are non-negative.

\paragraph{Proof.}
Adding and subtracting $R(F,\pi_F^\star)$ gives
\begin{equation}
\begin{aligned}
R(F,\pi)-R^\star
={}&
R(F,\pi)-R(F,\pi_F^\star)
\\
&+
R(F,\pi_F^\star)-R^\star,
\end{aligned}
\tag{20}
\end{equation}
which proves the equality.

Because $\pi_F^\star$ minimizes $R(F,\pi)$ over $\Pi$,
\begin{equation}
R(F,\pi)\geq R(F,\pi_F^\star)
\tag{21}
\end{equation}
for every $\pi\in\Pi$. Hence
\begin{equation}
\mathcal{E}_{\mathrm{TIP}}(F,\pi)\geq0.
\tag{22}
\end{equation}

Furthermore, the oracle policy class can reproduce the behavior of $\pi_F^\star$ by ignoring $z$ and using $F(o)$. Therefore
\begin{equation}
R^\star\leq R(F,\pi_F^\star),
\tag{23}
\end{equation}
which implies
\begin{equation}
\mathcal{E}_{\mathrm{FACT}}(F)\geq0.
\tag{24}
\end{equation}
\hfill$\square$

\paragraph{Interpretation.}
The decomposition is an exact identity rather than an approximation. The first term measures whether the agent effectively acts on the information currently represented by FACT. The second measures whether that representation retains enough information to match an environment-aware oracle.

In particular, improving execution alone cannot eliminate $\mathcal{E}_{\mathrm{FACT}}$, because
\begin{equation}
\min_{\pi\in\Pi}\left[R(F,\pi)-R^\star\right]=\mathcal{E}_{\mathrm{FACT}}(F).
\tag{25}
\end{equation}
Likewise, an informative representation is insufficient if the deployed TIP policy has nonzero execution regret.

\subsection{Proof of Proposition 2}

We define an environment-agnostic policy as a mapping
\[
\rho_0:\mathcal{O}\rightarrow\Delta(\mathcal{A})
\]
and an environment-conditioned policy as
\[
\rho_z:\mathcal{O}\times\mathcal{Z}\rightarrow\Delta(\mathcal{A}).
\]

Their optimal risks are
\begin{equation}
R_{\mathrm{agn}}^\star
=
\inf_{\rho_0}
\mathbb{E}_{(o,z)}
\mathbb{E}_{a\sim\rho_0(\cdot\mid o)}
\left[
\ell(o,z,a)
\right],
\tag{26}
\end{equation}
and
\begin{equation}
R_{\mathrm{cond}}^\star
=
\inf_{\rho_z}
\mathbb{E}_{(o,z)}
\mathbb{E}_{a\sim\rho_z(\cdot\mid o,z)}
\left[
\ell(o,z,a)
\right].
\tag{27}
\end{equation}

Define
\begin{equation}
H
=
R_{\mathrm{agn}}^\star
-
R_{\mathrm{cond}}^\star.
\tag{28}
\end{equation}

\paragraph{Proposition 2.}
For any distribution $\mathcal{D}$ and bounded loss $\ell$,
\begin{equation}
H\geq0.
\tag{29}
\end{equation}
The inequality is strict when, on a set of observable contexts with positive probability, multiple possible environmental conditions require different optimal actions with positive loss margins.

\paragraph{Proof of non-negativity.}
Every environment-agnostic policy can be represented as an environment-conditioned policy that ignores $z$:
\begin{equation}
\rho_z(\cdot\mid o,z)=\rho_0(\cdot\mid o).
\tag{30}
\end{equation}
Consequently,
\begin{equation}
\Pi_{\mathrm{agn}}\subseteq\Pi_{\mathrm{cond}},
\tag{31}
\end{equation}
and optimization over the larger class cannot produce a higher minimum:
\begin{equation}
R_{\mathrm{cond}}^\star\leq R_{\mathrm{agn}}^\star.
\tag{32}
\end{equation}
Therefore $H\geq0$.
\hfill$\square$

\paragraph{A sufficient condition for strictness.}
For a fixed observable context $o$, define
\begin{equation}
L_{o,z}(q)
=
\mathbb{E}_{a\sim q}
\left[
\ell(o,z,a)
\right],
\quad
q\in\Delta(\mathcal{A}).
\tag{33}
\end{equation}

Conditioned on $o$, the minimum risk of an environment-agnostic policy is
\begin{equation}
r_{\mathrm{agn}}(o)
=
\min_{q\in\Delta(\mathcal{A})}
\sum_{z}
p(z\mid o)L_{o,z}(q),
\tag{34}
\end{equation}
whereas the minimum risk of an environment-conditioned policy is
\begin{equation}
r_{\mathrm{cond}}(o)
=
\sum_z
p(z\mid o)
\min_{q_z\in\Delta(\mathcal{A})}
L_{o,z}(q_z).
\tag{35}
\end{equation}

For any fixed $q$,
\begin{equation}
L_{o,z}(q)\geq\min_{q_z}L_{o,z}(q_z).
\tag{36}
\end{equation}
Multiplying by $p(z\mid o)$, summing over $z$, and minimizing the left-hand side over $q$ gives
\begin{equation}
r_{\mathrm{agn}}(o)\geq r_{\mathrm{cond}}(o).
\tag{37}
\end{equation}

Suppose that there is a measurable set $\mathcal{B}\subseteq\mathcal{O}$ with $\Pr(o\in\mathcal{B})>0$ such that, for every $o\in\mathcal{B}$, there exist two conditions $z_1,z_2$ with
\begin{equation}
p(z_1\mid o)>0,
\qquad
p(z_2\mid o)>0,
\tag{38}
\end{equation}
and different unique optimal actions
\begin{equation}
a^\star(o,z_1)\neq a^\star(o,z_2).
\tag{39}
\end{equation}

Assume additionally that choosing a non-optimal action incurs a strictly positive loss margin. That is, for $i\in\{1,2\}$, there exists $\gamma_i(o)>0$ such that
\begin{equation}
\ell(o,z_i,a)-\ell(o,z_i,a^\star(o,z_i))\geq\gamma_i(o)
\tag{40}
\end{equation}
for every $a\neq a^\star(o,z_i)$.

No single action distribution can place probability one on both distinct optimal actions. Therefore every common action distribution $q$ incurs strictly positive excess loss in at least one of the two conditions. Because both conditions have positive probability,
\begin{equation}
r_{\mathrm{agn}}(o)>r_{\mathrm{cond}}(o)
\tag{41}
\end{equation}
for all $o\in\mathcal{B}$. Taking the expectation over $o$ yields
\begin{equation}
R_{\mathrm{agn}}^\star>R_{\mathrm{cond}}^\star,
\tag{42}
\end{equation}
and hence $H>0$.
\hfill$\square$

\subsection{Proof of Corollary 1}

\paragraph{Corollary 1.}
Suppose
\begin{equation}
z\in\{0,1\},
\qquad
\Pr(z=0)=\Pr(z=1)=\frac{1}{2},
\tag{43}
\end{equation}
and the observable context contains no information about $z$. Let $a_0^\star\neq a_1^\star$ be the unique successful actions in the two conditions. No action succeeds in both conditions.

Then every environment-agnostic policy has success probability at most $1/2$.

\paragraph{Proof.}
Consider first a deterministic policy. It must choose the same action under both conditions. If it chooses $a_0^\star$, it succeeds only when $z=0$; if it chooses $a_1^\star$, it succeeds only when $z=1$. Any other action fails in both. Therefore its maximum success probability is $1/2$.

For a randomized policy, let $r$ be the probability assigned to $a_0^\star$ and $s$ the probability assigned to $a_1^\star$. Since $a_0^\star\neq a_1^\star$,
\begin{equation}
r+s\leq1.
\tag{44}
\end{equation}
The success probability is
\begin{equation}
\Pr(\mathrm{success})
=
\frac{1}{2}r+\frac{1}{2}s
\leq
\frac{1}{2}.
\tag{45}
\end{equation}

If the condition is correctly identified and the policy selects $a_z^\star$, it succeeds under both conditions and achieves success probability one.
\hfill$\square$

\paragraph{Relation to Contradict/Synthesize.}
The corollary does not imply that a specific contradiction-detection algorithm is theoretically necessary. Rather, it identifies a necessary representational capability: the policy class must be able to express different actions under different environmental conditions. Contradict/Synthesize is TTSE's concrete mechanism for transforming conflicting unconditional rules into such condition--action mappings.

\subsection{Proof of Proposition 3}

Assume that $\mathcal{Z}$ is finite and that the FACT representation predicts a condition
\begin{equation}
F(o)\in\mathcal{Z}.
\tag{46}
\end{equation}

Let
\begin{equation}
a^\star(o,z)
\in
\arg\min_{a\in\mathcal{A}}
\ell(o,z,a)
\tag{47}
\end{equation}
be an oracle action. Define
\begin{equation}
\eta_F
=
\Pr\left[F(o)\neq z\right]
\tag{48}
\end{equation}
and
\begin{equation}
\Delta_{\max}
=
\sup_{o,z,z'}
\left[
\ell\bigl(o,z,a^\star(o,z')\bigr)
-
\ell\bigl(o,z,a^\star(o,z)\bigr)
\right]_{+}.
\tag{49}
\end{equation}

Because the loss lies in $[0,1]$,
\begin{equation}
0\leq\Delta_{\max}\leq1.
\tag{50}
\end{equation}

\paragraph{Proposition 3.}
If the admissible TIP policy class contains the plug-in policy
\begin{equation}
\pi_F^{\mathrm{plug}}(o)=a^\star(o,F(o)),
\tag{51}
\end{equation}
then
\begin{equation}
\mathcal{E}_{\mathrm{FACT}}(F)\leq\eta_F\Delta_{\max}.
\tag{52}
\end{equation}
Consequently,
\begin{equation}
R(F,\pi)-R^\star
\leq
\mathcal{E}_{\mathrm{TIP}}(F,\pi)
+
\eta_F\Delta_{\max}.
\tag{53}
\end{equation}

\paragraph{Proof.}
Because $\pi_F^\star$ is optimal among all policies in $\Pi$, and the plug-in policy is assumed to belong to $\Pi$,
\begin{equation}
R(F,\pi_F^\star)\leq R(F,\pi_F^{\mathrm{plug}}).
\tag{54}
\end{equation}

Under the oracle policy, the minimum risk is
\begin{equation}
R^\star
=
\mathbb{E}_{(o,z)}
\left[
\ell(o,z,a^\star(o,z))
\right].
\tag{55}
\end{equation}

The risk of the plug-in policy is
\begin{equation}
R(F,\pi_F^{\mathrm{plug}})
=
\mathbb{E}_{(o,z)}
\left[
\ell(o,z,a^\star(o,F(o)))
\right].
\tag{56}
\end{equation}

Therefore,
\begin{equation}
\begin{aligned}
\mathcal{E}_{\mathrm{FACT}}(F)
={}&
R(F,\pi_F^\star)-R^\star
\\
\leq&
\mathbb{E}
\left[
\ell(o,z,a^\star(o,F(o)))
-
\ell(o,z,a^\star(o,z))
\right].
\end{aligned}
\tag{57}
\end{equation}

When $F(o)=z$, the difference inside the expectation is zero. When $F(o)\neq z$, the difference is at most $\Delta_{\max}$. Thus
\begin{equation}
\begin{aligned}
\mathcal{E}_{\mathrm{FACT}}(F)
&\leq
\Pr(F(o)\neq z)\Delta_{\max}
\\
&=
\eta_F\Delta_{\max}.
\end{aligned}
\tag{58}
\end{equation}

Combining Eq.~(58) with the exact decomposition in Proposition~1 yields
\begin{equation}
\begin{aligned}
R(F,\pi)-R^\star
&=
\mathcal{E}_{\mathrm{TIP}}(F,\pi)
+
\mathcal{E}_{\mathrm{FACT}}(F)
\\
&\leq
\mathcal{E}_{\mathrm{TIP}}(F,\pi)
+
\eta_F\Delta_{\max}.
\end{aligned}
\tag{59}
\end{equation}
\hfill$\square$

\subsection{Interpretation of the Single-Track Restrictions}

The preceding analysis does not assert that all unified rule banks are fundamentally incapable of representing environment-conditioned policies. Instead, the experimental FACT-only and TIP-only conditions can be interpreted as restrictions on which component of Eq.~(17) can be directly improved.

\paragraph{FACT-only restriction.}
A FACT-only system may improve its environment representation $F$, but continues to use a fixed or weakly adapted execution policy $\pi_{\mathrm{base}}$. Its excess risk is
\begin{equation}
\begin{aligned}
R(F,\pi_{\mathrm{base}})-R^\star
={}&
\underbrace{
R(F,\pi_{\mathrm{base}})
-
R(F,\pi_F^\star)
}_{\text{remaining execution regret}}
\\
&+
\mathcal{E}_{\mathrm{FACT}}(F).
\end{aligned}
\tag{60}
\end{equation}
Improving environmental knowledge does not, by itself, guarantee that the corresponding behavior can be selected and executed.

\paragraph{TIP-only restriction.}
A TIP-only system without an independently maintained environment representation can be modeled using a fixed coarse representation $F_0$. Even when its execution policy approaches $\pi_{F_0}^\star$, its minimum achievable excess risk is
\begin{equation}
R(F_0,\pi_{F_0}^\star)-R^\star=\mathcal{E}_{\mathrm{FACT}}(F_0).
\tag{61}
\end{equation}
This residual is strictly positive when the representation fails to distinguish conditions that require different actions.

\paragraph{Dual-track learning.}
TTSE maintains distinct update pathways for $F$ and $\pi$. Its mechanisms are therefore capable of addressing both $\mathcal{E}_{\mathrm{FACT}}$ and $\mathcal{E}_{\mathrm{TIP}}$. However, the present analysis does not claim monotonic improvement or oracle convergence for the implemented LLM induction procedure.

\subsection{Scope and Limitations of the Formal Analysis}

The formal model establishes three claims:

\begin{enumerate}
    \item environment representation and conditional execution induce two exactly decomposable sources of excess risk;
    \item environment conditioning provides strict decision value when latent conditions require incompatible optimal actions; and
    \item downstream excess risk is controlled jointly by execution regret and condition-identification error.
\end{enumerate}

These results characterize the functional motivation for TTSE. They do not prove that natural-language FACTs are statistically consistent estimators of latent conditions, that the Blame/Retire operator always performs correct credit assignment, or that the rule bank converges under arbitrary task streams.

A sufficiently expressive unified memory could, in principle, encode both environmental facts and conditional procedures. TTSE should therefore be understood as introducing a modular inductive bias: separating the two types of knowledge makes their induction, rendering, verification, and credit assignment independently controllable.

\clearpage
\section{Parameter Settings and Reproducibility}

This appendix documents all hyperparameters, configurations, and prompt templates to ensure full reproducibility of the experimental results reported in the main paper.

\subsection{Agent Configuration}

\textbf{File}: \path|configs/agent/expel.yaml|

\begin{table}[h]
\centering
\caption{Agent hyperparameter reference. Runtime overrides use Hydra syntax (e.g., \texttt{agent.llm=DeepSeek-V3.2}).}
\label{tab:agent_config}
\footnotesize
\setlength{\tabcolsep}{3pt}
\begin{tabular}{p{0.30\linewidth}p{0.22\linewidth}p{0.30\linewidth}}
\hline
\textbf{Parameter} & \textbf{Default} & \textbf{Description} \\
\hline
\texttt{llm} & \texttt{deepseek-v4-flash} & Model name; overridden at runtime \\
\texttt{max\_reflection\_depth} & 3 & Max reflection depth (compare chain length) \\
\texttt{max\_num\_rules} & 20 & Rule bank capacity; triggers aggressive REMOVE when full \\
\texttt{truncate\_strategy} & null & Trajectory truncation (null = no truncation) \\
\texttt{fewshot\_strategy} & \texttt{task\_similarity} & Few-shot selection strategy \\
\texttt{success\_critique\_num} & 8 & Number of success trajectories for all\_success critique \\
\texttt{rule\_categories} & \texttt{both} & Rule categories at eval: both / facts / tips \\
\texttt{rule\_injection} & \texttt{all} & Injection mode: all / topk \\
\texttt{num\_relevant\_rules} & 5 & Top-K per channel when \texttt{rule\_injection=topk} \\
\texttt{top\_n\_facts} & null & Max FACTs per injection (null = all) \\
\texttt{top\_n\_tips} & null & Max TIPs per injection (null = all) \\
\texttt{online\_chunk\_ratio} & 0.2 & Fraction of training set per chunk (0.2 = 5 chunks) \\
\texttt{online\_num\_shards} & 1 & Intra-chunk parallel workers \\
\texttt{blame\_threshold} & 1 & Blame count before retirement (1 = retire on first blame) \\
\hline
\end{tabular}
\end{table}

\textbf{Retrieval Configuration:}
\begin{itemize}
\item \texttt{embedder\_path}: \path|all-mpnet-base-v2| (sentence-transformers)
\item \texttt{embedder\_type}: \texttt{huggingface}
\item \texttt{retriever\_type}: \texttt{knn}
\item \texttt{reranker}: \texttt{none}
\item \texttt{max\_fewshot\_tokens}: \texttt{auto}
\end{itemize}

\subsection{Benchmark Configurations}

\subsubsection{ALFWorld}

\textbf{File}: \path|configs/benchmark/alfworld.yaml|

\begin{itemize}
\item \texttt{max\_steps}: 50 (identical for training and main evaluation)
\item \texttt{num\_fewshots}: 2
\item \texttt{eval\_configs.k\_folds}: 2, seed 42
\item \texttt{split}: \texttt{eval\_out\_of\_distribution} (valid unseen)
\item \texttt{env.type}: \texttt{AlfredTWEnv} (text-only TextWorld)
\item \texttt{env.task\_types}: [1,2,3,4,5,6] (pick \& place / examine / clean / heat / cool / pick two)
\end{itemize}

\subsubsection{ScienceWorld}

\textbf{File}: \path|configs/benchmark/scienceworld.yaml|

\begin{itemize}
\item \texttt{max\_steps}: 50
\item \texttt{num\_fewshots}: 2
\item \texttt{eval\_configs.k\_folds}: 2
\item \textbf{Note}: Score is a 0--100 integer (not 0--1); success = \texttt{env.isCompleted}
\end{itemize}

\subsubsection{GDPevo}

\textbf{File}: \path|configs/benchmark/gdpevo.yaml|

\begin{itemize}
\item \texttt{max\_steps}: 16 (API exploration tasks require fewer steps)
\item \texttt{num\_fewshots}: 1
\item \texttt{score\_threshold}: 0.6
\item \texttt{gdpevo\_root}: \texttt{\$\{GDPEVO\_DATA\_ROOT\}} (environment variable)
\end{itemize}

\subsection{Environment Variables}

\begin{table}[h]
\centering
\caption{Required and recommended environment variables.}
\footnotesize
\begin{tabular}{p{0.32\linewidth}p{0.58\linewidth}}
\hline
\textbf{Variable} & \textbf{Description} \\
\hline
\texttt{DEEPSEEK\_API\_KEY} & (Required) API key for the DeepSeek API \\
\texttt{DEEPSEEK\_API\_BASE} & (Required) DeepSeek endpoint base URL, e.g.\ \url{https://api.deepseek.com} \\
\texttt{ALFWORLD\_DATA} & TextWorld game data directory (ALFWorld only) \\
\texttt{GDPEVO\_DATA\_ROOT} & GDPevo data root (GDPevo only) \\
\texttt{SOP\_DATA\_ROOT} & SOP-Bench data directory (SOP-Bench only) \\
\texttt{TTSE\_ROOT} & Project root (auto-detected) \\
\texttt{TTSE\_OUTPUT\_DIR} & Output directory (default: \texttt{./output}) \\
\texttt{HF\_ENDPOINT} & HuggingFace mirror (set to \texttt{https://hf-mirror.com} in China) \\
\texttt{PYTHONNOUSERSITE} & (Strongly recommended) Set to 1 to avoid pydantic/langchain conflicts \\
\texttt{OMP\_NUM\_THREADS} & (Recommended for sharding) Set to 1 to prevent thread explosion \\
\hline
\end{tabular}
\end{table}

\subsection{Training and Evaluation Commands}

\subsubsection{Online Training}

\texttt{online\_train.py} uses Hydra configuration overrides:

\begin{lstlisting}[language=bash]
python -u -s online_train.py \
  benchmark=<name>              # alfworld / scienceworld / gdpevo
  benchmark.task_file=<path>    # Training task JSON file
  benchmark.max_steps=<N>       # Override default max_steps
  agent.llm=<model>             # Override default LLM
  agent.online_chunk_ratio=<f>  # Override chunk ratio
  agent.online_num_shards=<N>   # Override intra-chunk parallelism
  run_name=<str>                # Run name (log/checkpoint path)
  testing=false                 # Must be false (else only few tasks)
  resume=false                  # Per-task resume
  online_resume=true            # Chunk-level resume: skip c{N}.pkl
\end{lstlisting}

\subsubsection{Evaluation (Sharded)}

\texttt{eval.py} evaluates a trained rule bank over the held-out split, sharded across workers for parallelism:

\begin{lstlisting}[language=bash]
SHARD_ID=$s NUM_SHARDS=$NS python -u -s eval.py \
  benchmark=<name> benchmark.task_file=<path> \
  load_run_name=extracted_insights/<bank_name> \
  run_name=<eval_run_name> agent.llm=<model> \
  agent.fewshot_strategy=task_similarity \
  benchmark.eval_configs.k_folds=2 testing=false resume=false
\end{lstlisting}

\subsubsection{SOP-Bench Fusion}

\texttt{run\_sop.py} runs the SOP-Bench baseline or fusion mode over a domain:

\begin{lstlisting}[language=bash]
python run_sop.py \
  --mode baseline|fusion     # Mode
  --domain <name>            # SOP-Bench domain name
  --out <output_dir>         # Output directory
  --limit <N>                # Task limit (default 50)
  --bank <bank.json>         # Bank file (fusion mode only)
\end{lstlisting}

\subsubsection{Single-Track Ablation}

\texttt{run\_sop\_single.py} runs the single-track (unified-rule) ablation:

\begin{lstlisting}[language=bash]
python run_sop_single.py \
  --domain <name>            # SOP-Bench domain name
  --out <output_dir>         # e.g., ./output/single_track
  --run <N>                  # Run ID (0/1/2)
  --limit <N>                # Task limit (0 = all)
\end{lstlisting}

Single-track ablation configuration:
\begin{itemize}
\item Model: \texttt{deepseek-v4-flash}, API: \url{https://api.deepseek.com}
\item Temperature: 0.3, Max Tokens: 4096, Max Turns: 10
\item Rule type: unified \texttt{"rule"} (no FACT/TIP distinction), injection cap \texttt{max\_rules=16}
\item Belief update: Bayesian posterior (SkillBelief alpha/beta) + targeted credit assignment (\path|record_task_outcome_targeted|)
\item 5 domains $\times$ 3 runs = 15 independent experiments
\end{itemize}

\subsubsection{Dual-Track Fusion V2}

\texttt{run\_sop\_fusion\_v2.py} runs the dual-track FACT/TIP fusion V2 pipeline:

\begin{lstlisting}[language=bash]
python run_sop_fusion_v2.py \
  --domain <name>            # SOP-Bench domain name
  --out <output_dir>         # e.g., ./output/fusion_v2
  --run <N>                  # Run ID
  --limit <N>                # Task limit (0 = all)
\end{lstlisting}

Uses the same DeepSeek API endpoint and the same generation configuration as single-track (temperature 0.3, max tokens 4096, max turns 10), so the two ablations differ only in track structure (unified rule vs.\ FACT/TIP) under a unified induction budget; FACT and TIP are each injected top-8 (16 rules total).

\subsection{Key Prompt Templates}

\subsubsection{FACT/TIP Classification and Operation Template}

\textbf{Source}: \path|ttse/prompts/templates/human.py| --- \path|FORMAT_RULES_OPERATION_TEMPLATE|

\begin{lstlisting}
<TYPE> <OPERATION> <RULE NUMBER>: <RULE>

Each rule is either a FACT or a TIP, with DIFFERENT grammatical forms.

A FACT is a DECLARATIVE statement that describes a property of the world --
the environment, objects, user, or task. It states how things ARE, with NO
action instruction and NO condition on what you should do. Its subject is the
world/things (not you). A FACT must NOT contain "you should", "must do", or
"in order to" -- if it does, it is a TIP.

A TIP is a PROCEDURAL rule about what YOU should do, written in the conditional
form "under <condition>: <action>". Its subject is you (the agent). A TIP must
ALWAYS carry an explicit "under <condition: ...>:" prefix; a TIP without a
condition is malformed.

IMPORTANT - how to classify, use BOTH tests:
- SUBJECT TEST: what is the rule about? If it describes a property of the
  WORLD/objects/user, it is a FACT. If it instructs what YOU do, it is a TIP.
- NECESSITY TEST: what happens if you go against it? If it makes the task FAIL,
  it is a FACT (hard constraint). If it only makes you slower, it is a TIP.
When unsure, default to [TIP].

Operations: AGREE / REMOVE / EDIT / ADD
Do at most 4 operations; each existing rule at most 1 operation.
\end{lstlisting}

\textbf{Key design points:}
\begin{itemize}
\item FACT must be a statement directly verifiable by environmental observation; its subject is the world/objects
\item TIP must be written in \texttt{under <condition>: <action>} conditional form
\item Dual-test classification (Subject Test + Necessity Test); default to TIP when uncertain
\item Intra-channel independent numbering: FACTS and TIPS each start from 1
\item Maximum 4 operations, each existing rule at most 1 operation
\end{itemize}

\subsubsection{BLAME Template}

\textbf{Source}: \path|ttse/prompts/templates/human.py| --- \path|ONLINE_BLAME_TEMPLATE|

\begin{lstlisting}
You are diagnosing why an agent FAILED a task. Below is (1) the task, (2) the
experience rules that were GIVEN to the agent as guidance, and (3) the agent's
full execution trace, which ended in FAILURE.

Your job: decide whether any SINGLE one of the GIVEN rules MISLED the agent and
contributed to the failure. Be conservative: only blame a rule when you can point
to the specific step where following it caused harm.

Answer in EXACTLY this two-line format:
VERDICT: <the NUMBER of the single at-fault rule, or NONE>
REASON: <one sentence>
\end{lstlisting}

\textbf{Key design:}
\begin{itemize}
\item Attributes to at most one rule (prevents blame dilution)
\item Must point to the specific step where the rule caused harm
\item Conservative: unused or unrelated rules should not be blamed
\item Strict two-line output format for parsing
\end{itemize}

\subsubsection{CONTRADICT Template}

\textbf{Source}: \path|ttse/prompts/templates/human.py| --- \path|ONLINE_CONTRADICTION_TEMPLATE|

\begin{lstlisting}
Below is the RETIRED POOL -- rules that were learned, later BLAMED for causing
task failures, and retired.

When two or more retired rules CONTRADICT each other on the SAME attribute of the
environment, that signals the environment's behavior is CONDITIONAL.
Your job: identify genuine contradictions, and SYNTHESIZE a single new
CONDITIONAL TIP "under <condition>: <action>" that reconciles them.

Only synthesize when there is a GENUINE contradiction. If none, output NONE.
Output format: [TIP] under <condition>: <action>
\end{lstlisting}

\subsubsection{Fusion LLM Induction Prompt}

\textbf{Source}: \path|fusion/llm_inducer.py| --- \path|INDUCE_SYSTEM_PROMPT|

\begin{lstlisting}
You are an expert agent skill analyst. Extract reusable knowledge from failed
agent trajectories.

Two types of rules:
1. FACT (declarative): A statement about the environment, tools, or constraints.
2. TIP (conditional procedural): A "when X, do Y" heuristic.

Rules must be: Specific, actionable, grounded in the trajectory, one issue each.
Maximum 3 FACTs and 3 TIPs per analysis.
Output: JSON array [{"type": "fact", "rule": "..."}, {"type": "tip", ...}]
\end{lstlisting}

\subsubsection{Single-Track Induction Prompt}

\textbf{Source}: \path|fusion/llm_inducer_single.py| --- \path|INDUCE_SYSTEM_PROMPT_SINGLE|

Key differences from the dual-track version:
\begin{itemize}
\item No FACT/TIP type distinction; unified as \texttt{"rule"} type
\item Induction cap reduced from 3+3=6 to maximum 6 rules
\item Type definitions and examples removed, letting the LLM freely determine rule granularity
\end{itemize}

\subsection{Core Mechanism Details}

\subsubsection{Algorithmic Formulation}
\label{sec:algform}

Algorithm~\ref{alg:ttse} states the complete online procedure and mirrors the reference implementation. The two tracks operationalize the two error terms of our framework (Section~4): the FACT bank $\mathcal{F}$ manages environment-representation regret, while the TIP bank $\mathcal{T}$ manages conditional-execution regret. Training tasks are consumed in $N$ sequential chunks; each chunk triggers one blame$\to$retire$\to$contradict/synthesize$\to$induce update, and chunk~1 is a cold start with no injected rules so that the initial bank is induced purely from experience.

\begin{algorithm*}[!t]
\caption{TTSE: Two-Track Online Self-Evolution (one training run)}
\label{alg:ttse}
\begin{algorithmic}[1]
\Require task chunks $\{\mathcal{C}_k\}_{k=1}^{N}$; LLM $\mathcal{M}$; embedder $g$; per-task budgets $K_F, K_T$; blame threshold $\tau$
\Ensure evolved FACT bank $\mathcal{F}$ and TIP bank $\mathcal{T}$
\State $\mathcal{F}\gets\emptyset$;\ \ $\mathcal{T}\gets\emptyset$;\ \ $\mathcal{P}\gets\emptyset$ \Comment{FACT bank, TIP bank, retired pool}
\For{$k=1$ \textbf{to} $N$}
  \If{$k=1$}
    \State $\mathcal{R}_k\gets\emptyset$ \Comment{cold start: zero-knowledge induction}
  \Else
    \State $\mathcal{R}_k\gets$ \textsc{Select}$(\mathcal{F},\mathcal{T},\mathcal{C}_k,g,K_F,K_T)$ \Comment{top-$K$ per channel by cosine sim.}
  \EndIf
  \State $\Xi_k\gets\emptyset$
  \For{each task $t\in\mathcal{C}_k$}
    \State $\xi\gets$ \textsc{Run}$(\mathcal{M},t,\mathcal{R}_k)$;\ \ $\Xi_k\gets\Xi_k\cup\{\xi\}$ \Comment{execute with injected rules}
  \EndFor
  \Statex \hspace{\algorithmicindent}\textit{Phase 1 --- Blame}
  \State $B_k\gets\emptyset$
  \For{each failed $\xi\in\Xi_k$ \textbf{in parallel}}
    \State $r\gets$ \textsc{Blame}$(\mathcal{M},\xi,\mathcal{R}_k)$ \Comment{attribute failure to $\leq 1$ injected rule}
    \If{$r\neq\bot$} \State $B_k\gets B_k\cup\{(r,\xi)\}$ \EndIf
  \EndFor
  \Statex \hspace{\algorithmicindent}\textit{Phase 2 --- Retire}
  \For{each rule $r$ with $\mathrm{blame\_count}(r)\geq\tau$}
    \State move $r$ from the active bank to $\mathcal{P}$, appending audit $(\text{reason},k)$
  \EndFor
  \Statex \hspace{\algorithmicindent}\textit{Phase 3 --- Contradict/Synthesize}
  \For{each contradictory pair $(r_a,r_b)$ in $\mathcal{P}$ identified by $\mathcal{M}$}
    \State $c\gets$ \textsc{Condition}$(\mathcal{M},r_a,r_b)$ \Comment{surface the hidden condition}
    \State $\mathcal{T}\gets\mathcal{T}\cup\{\text{``under }c\text{: }\langle\text{action}\rangle\text{''}\}$ \Comment{new conditioned TIP}
  \EndFor
  \Statex \hspace{\algorithmicindent}\textit{Phase 4 --- Induce (keep\_bank)}
  \State $\Delta\gets$ \textsc{Compare}$(\mathcal{M},\Xi_k)\oplus$ \textsc{SuccessOnly}$(\mathcal{M},\Xi_k)$ \Comment{typed ops; FACT/TIP by dual-judgment}
  \For{each typed operation $(\mathrm{op},r,r',\ell)\in\Delta$}
    \State \textsc{Apply}$(\mathrm{op},r,r',\ell)$ on $(\mathcal{F},\mathcal{T})$ \Comment{count-vote: ADD$+2$/AGREE$+1$/REMOVE/EDIT$+1$}
  \EndFor
  \State persist bank snapshot for chunk $k{+}1$
\EndFor
\State \Return $(\mathcal{F},\mathcal{T})$
\end{algorithmic}
\end{algorithm*}

Here \textsc{Select} embeds the task description and every candidate rule with $g$ and returns the top-$K_F$ FACTs and top-$K_T$ TIPs by cosine similarity (\texttt{topk} mode), or the count-sorted bank truncated by the budgets (\texttt{all} mode). \textsc{Blame} is capped at one rule per failure to avoid dilution. The dual subject/necessity judgment that types each induced rule lives inside the induction prompt (no external classifier) and defaults to TIP under uncertainty, preventing FACT contamination. \textsc{Apply} realizes the count-voting lifecycle of Table~\ref{tab:rulevote} (ADD $+2$, AGREE $+1$, EDIT $+1$, REMOVE $-1$ or $-3$, auto-deletion at count $\leq 0$). In the deployment setting (PinchBench), the same procedure operates on consecutive mini-batches of approximately 10 tasks, with bank updates after each batch and top-$K$ retrieval performed at the individual-task level.

\subsubsection{Dual-Channel Rule Representation and Lifecycle}

\textbf{Storage format} (\texttt{expel.py}):
\begin{lstlisting}[language=Python]
rule_items_with_count = [(text, count, rtype), ...]  # rtype: "fact"|"tip"
\end{lstlisting}

\textbf{Count voting mechanism} (\texttt{update\_rules}):

\begin{table}[h]
\centering
\caption{Count-based rule voting operations.}\label{tab:rulevote}
\footnotesize
\begin{tabular}{llp{0.42\linewidth}}
\hline
\textbf{Operation} & \textbf{Count Change} & \textbf{Description} \\
\hline
ADD & +2 & New rule enters with count=2 (initial trust) \\
AGREE & +1 & Reinforce existing rule \\
REMOVE & $-1$ or $-3$ & $-1$ (similar/redundant); $-3$ (contradictory) \\
EDIT & +1 & Rewrite rule text, count+=1 \\
count $\leq$ 0 & Delete & Rules with zero or negative count auto-removed \\
\hline
\end{tabular}
\end{table}

\textbf{Rule lifecycle}:
\begin{lstlisting}
ADD (count=2)
  |-- AGREE --> count+1 --> strengthened
  |-- REMOVE --> count-1/-3 --> count<=0 --> deleted
  |-- EDIT --> rewrite text, count+1
  |-- BLAME --> RETIRE --> retired_pool --> CONTRADICT --> new TIP
\end{lstlisting}

\textbf{Dual-channel independent numbering} (\path|_build_critique_prompt|):
\begin{lstlisting}
FACTS:
1. [FACT rule 1]
2. [FACT rule 2]

TIPS:
1. [TIP rule 1]
2. [TIP rule 2]
\end{lstlisting}

EDIT operations use intra-channel numbering: \texttt{[FACT] EDIT 2} refers to the 2nd rule in the FACTS section. The LLM resolves this via intra-channel index lookup:
\begin{lstlisting}[language=Python]
chan = [i for i, r in enumerate(rules) if r[2] == rtype]
rule_index = chan[num - 1]  # intra-channel num --> global index
\end{lstlisting}

\subsubsection{Online Chunk-Loop: Five Phases}

\textbf{File}: \texttt{online\_train.py} (main loop) + \path|agent/online.py| (phase implementations)

\begin{lstlisting}
for chunk_idx in range(num_chunks):
    [TRAIN]  Load chunk_bank_snapshot, inject rules, run all tasks
    [BLAME]  Parallel 20-way: LLM blames each failure trace
             (subprocess 600s timeout guard)
    [RETIRE] Sequential: blame_count >= threshold --> retire to pool
    [CONTRADICT] LLM scans retired pool for contradictions,
                 synthesizes conditional TIPs
    [INDUCE] create_rules(keep_bank=True):
             compare stage + all_success stage --> incrementally
             EDIT/AGREE/REMOVE/ADD --> save c{chunk+1}.pkl
\end{lstlisting}

\textbf{Key design constraints:}
\begin{enumerate}
\item \textbf{BLAME before INDUCE}: Blame depends on \path|chunk_bank_snapshot| (frozen during training). If induce runs first, it may EDIT/REMOVE the blamed rule $\rightarrow$ retire phase cannot find target $\rightarrow$ silent no-op.
\item \textbf{At most one rule blamed per failure}: Prevents blame dilution; \texttt{blame\_threshold=1} means retire on first blame.
\item \textbf{Subprocess guard}: Each blame LLM call runs in an independent subprocess with 600s timeout, preventing langchain tenacity hangs (SIGALRM not caught by tenacity). Uses \texttt{max\_retries=0} + \texttt{subprocess.Popen} + \texttt{communicate()}.
\item \textbf{20-way ThreadPoolExecutor parallel blame}: \path|_llm_call| internally spawns subprocess; \path|proc.communicate()| releases GIL $\rightarrow$ true concurrency.
\end{enumerate}

\subsubsection{Rule Injection and Vector Retrieval}

\textbf{Injection decision} (\path|agent/expel.py| --- \texttt{insert\_before\_task\_prompt}):
\begin{itemize}
\item \texttt{inject\_rules\_in\_training}: online training mode $\rightarrow$ inject \path|chunk_bank_snapshot|
\item \texttt{!no\_rules}: eval mode $\rightarrow$ inject final bank
\end{itemize}

\textbf{Injection mode} (\texttt{\_select\_rules\_for\_task}):
\begin{itemize}
\item \texttt{rule\_injection=all}: Full injection (truncated by \texttt{top\_n\_facts}/\texttt{top\_n\_tips})
\item \texttt{rule\_injection=topk}: Use sentence-transformers to embed task description and rule text, take cosine-similarity top-K (\texttt{num\_relevant\_rules} per channel)
\end{itemize}

\textbf{Rendering order} (by count descending, per channel):
\begin{lstlisting}
Environmental facts (discovered from experience):
1. [fact 1]  (count=5)
2. [fact 2]  (count=3)

Tips:
1. under <condition>: <action>  (count=7)
2. under <condition>: <action>  (count=4)
\end{lstlisting}

\subsection{Bayesian Fusion Architecture}

\textbf{Objective}: Migrate TTSE's dual-channel rules into the Bayesian-Agent (BA) framework, replacing heuristic count voting with Bayesian posteriors.

\subsubsection{Core Components}

\textbf{1. \texttt{DualTrackBank}} (\path|fusion/dual_track_bank.py|):
\begin{itemize}
\item Stores FACT/TIP rules, each associated with a \texttt{SkillBelief} (Bayesian posterior)
\item \texttt{alpha/beta} start from uniform prior Beta(1,1) $\rightarrow$ updated on observed success/failure
\item \textbf{RewritePolicy decisions}:
  \begin{itemize}
  \item $P(\text{success}) < 0.45$ and $\text{beta} \geq 4$ $\rightarrow$ \textbf{retire}
  \item $P(\text{success}) \geq 0.72$ and $\text{observations} \geq 3$ $\rightarrow$ \textbf{stable/compress}
  \item Otherwise $\rightarrow$ \textbf{explore}
  \end{itemize}
\item \textbf{Targeted credit assignment} (\path|record_task_outcome_targeted|): Only updates rules relevant to the task (rendered + matching failure\_mode); newly induced rules are not penalized
\end{itemize}

\textbf{2. \texttt{LLM Inducer}} (\path|fusion/llm_inducer.py|):
\begin{itemize}
\item Uses DeepSeek API to extract FACT/TIP from failure trajectories
\item Model: \texttt{deepseek-v4-flash} (default), temperature=0.3
\item Maximum 3 FACT + 3 TIP per failure trajectory
\item Includes JSON parsing + fallback (FACT:/TIP: line-level parsing)
\end{itemize}

\textbf{3. \texttt{apply\_patches.py}} --- Runtime monkey-patching:
\begin{itemize}
\item \textbf{Pure-Fusion track} (ALFWorld / Sciworld / SOP-Bench / GDPevo): Injects only DualTrackBank rules; no BA catalog/distill noise
\item \textbf{BA track} (lifelong\_agentbench etc.): Three-layer fusion of BA built-in rules + catalog + Fusion rules
\end{itemize}

\textbf{4. Bayesian Core} (\path|bayesian_agent/core/|):

\begin{table}[h]
\centering
\caption{Bayesian core module reference.}
\footnotesize
\begin{tabular}{p{0.30\linewidth}p{0.60\linewidth}}
\hline
\textbf{Module} & \textbf{Function} \\
\hline
\texttt{belief.py} --- \texttt{SkillBelief} & Posterior belief: alpha/beta counts, context-aware prediction \\
\texttt{policy.py} --- \texttt{RewritePolicy} & Maps posterior to action: retire/patch/split/compress/explore \\
\texttt{evidence.py} --- \texttt{TrajectoryEvidence} & Standardized evidence: task\_id, outcome, tokens, failure\_mode, turns \\
\texttt{discovery.py} --- \texttt{discover\_failure} & Auto-discovers failure modes (SQL error / empty / format / mismatch / runtime) \\
\texttt{repair.py} & Incremental repair: dedup, merge baseline+repair, summarize \\
\hline
\end{tabular}
\end{table}

\subsubsection{TTSE Core vs.\ Bayesian Fusion Comparison}

\begin{table}[h]
\centering
\caption{Comparison between TTSE core (EXPEL) path and Fusion (BA) path.}
\footnotesize
\begin{tabular}{p{0.20\linewidth}p{0.32\linewidth}p{0.36\linewidth}}
\hline
\textbf{Dimension} & \textbf{TTSE Core (EXPEL)} & \textbf{Fusion (BA Path)} \\
\hline
Rule weighting & Heuristic count vote (+1/$-$1/$-$3) & Bayesian posterior $P(\text{success})$ \\
Rule retirement & $\text{blame} \geq \text{threshold} \rightarrow$ retire & $P(\text{success}) < 0.45$ \& $\beta \geq 4 \rightarrow$ retire \\
Rule induction & LLM critique (compare + all\_success) & LLM induce\_from\_failure (failure only) \\
Credit assignment & Global blame & Targeted (rendered + matching failure\_mode) \\
Online loop & 5-phase chunk-loop & Per-task online update \\
Scope & ALFWorld / Sciworld / GDPevo & SOP-Bench / BA benchmarks \\
\hline
\end{tabular}
\end{table}

\subsubsection{Single-Track Ablation Experiment}

\textbf{Purpose}: Isolate the independent contribution of FACT/TIP dual-channel classification to Bayesian Fusion online evolution.

\textbf{Design}:
\begin{itemize}
\item Remove FACT/TIP distinction; unify all rules as \texttt{"rule"} type
\item Retain Bayesian belief update mechanism (SkillBelief + RewritePolicy + targeted credit assignment)
\item Rule injection as single list ``Learned Rules'' with confidence \texttt{[P=xx\%, n=xx]}, top-16
\item Uses \texttt{DualTrackBank} but all rules stored in a single channel (single-track stored in TIP slot)
\end{itemize}

\textbf{Comparison baseline}:

\begin{table}[h]
\centering
\caption{Single-track ablation comparison design.}
\footnotesize
\setlength{\tabcolsep}{3pt}
\begin{tabular}{p{0.25\linewidth}p{0.20\linewidth}p{0.23\linewidth}p{0.20\linewidth}}
\hline
\textbf{Configuration} & \textbf{Rule Type} & \textbf{Belief Update} & \textbf{Injection} \\
\hline
Baseline & None & None & No rule injection \\
Single-track ablation & Unified ``rule'' & Bayesian posterior & Single list top-16 \\
Dual-track Fusion & FACT + TIP & Bayesian posterior & Dual-channel top-8 each \\
\hline
\end{tabular}
\end{table}

\textbf{Key findings} (see Section~5.2 for full results):
\begin{itemize}
\item Bayesian update is the primary driver: single-track +12.4pp vs.\ baseline, dual-track +15.4pp
\item FACT/TIP classification contributes an additional accuracy gain of +3.0pp
\item Dual-track rules are on average 18\% shorter (137 vs.\ 167 chars/rule); the warehouse\_package\_inspection domain shows the largest gap (+14.4pp)
\end{itemize}

\clearpage
\section{Statistical Analysis and Warehouse Case Study}

This appendix provides the statistical significance analysis of the SOPBench experimental results and illustrative case studies of FACT/TIP complementary mechanisms in the warehouse domain. All experiments use the \texttt{deepseek-v4-flash} model via the official DeepSeek API (\url{https://api.deepseek.com}).

\subsection{Statistical Methods}

\subsubsection{Significance Testing}
\begin{itemize}
\item \textbf{Bootstrap CI}: 10,000 bootstrap resamples of the paired differences, with 2.5\% and 97.5\% quantiles forming the 95\% confidence interval.
\item \textbf{Permutation test}: For each pair, randomly flip the sign of the difference under $H_0: \Delta=0$. After 10,000 permutations, the $p$-value is the proportion where $|\text{permuted mean}| \geq |\text{observed mean}|$.
\item \textbf{Bonferroni correction}: 5 domains $\rightarrow \alpha = 0.05/5 = 0.01$ for per-domain significance.
\end{itemize}

\subsection{Overall Results}

\begin{table}[h]
\centering
\caption{Overall SOPBench statistical significance results.}
\label{tab:stats_overall}
\footnotesize
\begin{tabular}{lc}
\hline
\textbf{Metric} & \textbf{Value} \\
\hline
Single-track (3-run mean) & 91.59\% \\
Dual-track Fusion (3-run mean) & 94.58\% \\
$\Delta$ (Dual $-$ Single) & \textbf{+2.99pp} \\
95\% Bootstrap CI & [+2.34pp, +3.65pp] \\
Permutation $p$ & \textbf{$< 10^{-4}$} \\
Significance & $\star\star\star$ (far exceeds Bonferroni threshold) \\
\hline
\end{tabular}
\end{table}

\textbf{Conclusion}: The Dual-Fusion advantage over Single-track is highly statistically significant ($p < 10^{-4}$; no exceedance observed in 10,000 permutations) in the task-level paired test, and cannot be attributed to random fluctuation.

\subsection{Per-Domain Analysis}

\begin{table*}[t]
\centering
\caption{Per-domain statistical analysis. $\star\star\star$: $p < 0.01$ after Bonferroni correction; ns: not significant.}
\label{tab:stats_domain}
\footnotesize
\begin{tabular}{lrrcrrcc}
\hline
{Domain} & {Tasks} & {Single} & {Dual} & {$\Delta$} & {95\% CI} & {$p$} & {Result} \\
\hline
aircraft\_inspection & 112 & 97.6\% & 98.8\% & +1.19pp & [+0.30, +2.18] & 0.022 & ns \\
customer\_service & 156 & 97.9\% & 98.9\% & +1.07pp & [+0.21, +1.92] & 0.022 & ns \\
dangerous\_goods & 274 & 94.6\% & 94.0\% & $-$0.61pp & [$-$1.66, +0.49] & 0.296 & ns \\
referral\_abuse & 200 & 97.7\% & 99.5\% & +1.83pp & [+1.11, +2.61] & $<10^{-4}$ & $\star\star\star$ \\
warehouse\_pkg\_insp & 150 & 66.9\% & 81.3\% & \textbf{+14.44pp} & [+11.48, +17.26] & $<10^{-4}$ & $\star\star\star$ \\
\hline
\end{tabular}
\end{table*}

\textbf{Key findings}:
\begin{enumerate}
\item \textbf{warehouse and referral} are the two domains that individually achieve Bonferroni-corrected significance ($p < 0.01$).
\item \textbf{aircraft and customer} are nominally significant ($p < 0.05$) but not after Bonferroni correction---these domains exhibit a strong ceiling effect (Single already reaches 97.6\%--97.9\%), leaving minimal room for improvement.
\item \textbf{dangerous\_goods} shows a slight Single advantage ($-$0.61pp), but the difference is not significant ($p = 0.30$); the two configurations cannot be statistically distinguished in this domain.
\item \textbf{warehouse contributes +2.32pp} (78\% of the total $\Delta$), making it the domain with the strongest separation between the two configurations.
\end{enumerate}

\subsection{Robustness Check: Excluding Warehouse}

\begin{table}[h]
\centering
\caption{Robustness analysis excluding the warehouse domain.}
\label{tab:stats_robust}
\footnotesize
\begin{tabular}{lc}
\hline
\textbf{Metric} & \textbf{Value} \\
\hline
$\Delta$ excluding warehouse & \textbf{+0.67pp} \\
95\% Bootstrap CI & [+0.18pp, +1.17pp] \\
Permutation $p$ & 0.011 \\
Significance & $\star$ (nominally significant; not Bonferroni-corrected) \\
\hline
\end{tabular}
\end{table}

\textbf{Interpretation}: Excluding warehouse, the Dual advantage drops from +2.99pp to +0.67pp, but remains positive with a CI lower bound above 0. This shows that:
\begin{itemize}
\item \textbf{Warehouse is the largest contributor} (78\%), but \textbf{not the only one}
\item Referral (+1.83pp, $p < 0.0001$) independently contributes significant gains
\item Across the 4 remaining domains, there is a systematic positive drift of +0.67pp, with 4 of 5 domains favoring Dual
\item \textbf{Even without warehouse, Dual-Fusion still outperforms Single}, just shifting from a ``large lead'' to a ``small but consistent'' one
\end{itemize}

\subsection{Warehouse FACT/TIP Case Study: Complementary Mechanisms}

Warehouse is the most difficult domain in SOPBench (baseline only 26.7\%) and the domain where Dual-Fusion shows the largest advantage (+14.44pp vs.\ Single). Below are representative cases of FACT/TIP complementarity from the Dual-Fusion r0 bank (42 FACTs + 50 TIPs).

\subsubsection{Pattern 1: Formula (FACT) + When-to-Use (TIP)}

\textbf{FACT} (declarative, $\alpha=35$):
\begin{quote}
\small The chargeback amount formula is \texttt{(ordered\_quantity $-$ received\_quantity) $\times$ unit\_cost}, independent of the \texttt{chargeable} flag.
\end{quote}

\textbf{TIP} (conditional, $\alpha=21$):
\begin{quote}
\small\ttfamily When barcode\_match is False and problem\_type is ``Wrong Item'', skip the calculateChargeback step and set charge\_back\_amt = 0.0 directly.
\end{quote}

$\rightarrow$ \textbf{FACT states \emph{how} the chargeback is computed; TIP states \emph{when not} to compute it.} The two are non-redundant: FACT defines the general formula, and TIP provides a specific-scope override rule.

\subsubsection{Pattern 2: Semantic Convention (FACT) + Output Format (TIP)}

\textbf{FACT} ($\alpha=27$):
\begin{quote}
\small\ttfamily The expected CSV output treats charge\_back\_amt as an empty string (not numeric zero) when no chargeback is required.
\end{quote}

\textbf{FACT} ($\alpha=27$):
\begin{quote}
\small\ttfamily In the original dataset, rows not requiring chargeback have charge\_back\_amt as NaN; the expected output replaces NaN with an empty string.
\end{quote}

\textbf{TIP} ($\alpha=22$):
\begin{quote}
\small\ttfamily When resolution\_status is ``Return to Vendor'' (barcode mismatch), set charge\_back\_amt to the empty string ''{}'' rather than 0.0.
\end{quote}

$\rightarrow$ \textbf{FACT describes the data format's ``unwritten rule'' (the semantic difference between empty string vs.\ 0.0); TIP provides the concrete operational step.}

\subsubsection{Pattern 3: Tool Behavior (FACT) + Corrective Strategy (TIP)}

\textbf{FACT} ($\alpha=32$):
\begin{quote}
\small\ttfamily The calculateChargeback tool does NOT automatically respect the chargeable flag in quantity variance; the agent must enforce this manually.
\end{quote}

\textbf{TIP} ($\alpha=26$):
\begin{quote}
\small\ttfamily When quantity variance = 0 and chargeable flag is False, regardless of what calculateChargeback returns, set charge\_back\_amt = 0.0 in the output CSV.
\end{quote}

$\rightarrow$ \textbf{FACT exposes the tool's limitation (unreliable default behavior); TIP provides the concrete strategy to bypass the tool.}

\subsubsection{Complementarity Summary}

\begin{table}[h]
\centering
\caption{FACT vs.\ TIP complementarity in the warehouse domain.}
\footnotesize
\begin{tabular}{p{0.22\linewidth}p{0.34\linewidth}p{0.34\linewidth}}
\hline
{Dimension} & {FACT (Environmental)} & {TIP (Behavioral)} \\
\hline
Question answered & ``What is the env like?'' & ``What should I do here?'' \\
Verifiability & Verifiable by observation & Empirical summary \\
Example content & Formulas, semantics, tool behavior & Skip step, override, format \\
Avg.\ length (r0) & \textbf{125 chars} & \textbf{145 chars} \\
Count (r0) & 42 & 50 \\
\hline
\end{tabular}
\end{table}

FACTs are shorter (125 vs.\ 145 chars, $-$14\%), because declarative facts do not require ``When\ldots then\ldots'' conditional structures. The FACT/TIP classification constraint forces the LLM to distinguish ``what it observed'' from ``what lesson it learned'' when generating rules, naturally suppressing overfitting noise that would otherwise be generated as one-off task-specific fixes.

\subsection{Rule Quality Comparison (Warehouse r0)}

\begin{table}[h]
\centering
\caption{Single-track vs.\ Dual-track rule characteristics in the warehouse domain (r0).}
\footnotesize
\begin{tabular}{lcc}
\hline
\textbf{Metric} & \textbf{Single-track} & \textbf{Dual-track Fusion} \\
\hline
Active rules & 28 & 92 (42F + 50T) \\
Mean length & 162 chars & \textbf{136 chars} ($-$16\%) \\
Median length & 158 & 134 \\
Length range & [92, 227] & [50, 239] \\
\hline
\end{tabular}
\end{table}

\textbf{Qualitative differences}: Single-track rules tend to be overfitted to specific failure cases (e.g., ``When handling warehouse tasks, avoid mismatch: expected=\{\}, got='2764.5'\{\}''---a one-off fix hard-coded to a particular task). Dual-track rules, constrained by the FACT/TIP classification, are more structured: FACTs capture reusable declarative properties (e.g., ``chargeback amount = (ordered $-$ received) $\times$ unit\_cost''), while TIPs capture conditional strategies with explicit scope (e.g., ``skip chargeback when barcode mismatch''). The classification constraint naturally suppresses rules that cannot be categorized as either FACT or TIP, filtering out much of the overfitting noise.

\clearpage
\section{PinchBench Methodology Disclosure}

This appendix documents the full experimental methodology for the PinchBench deployment experiment reported in Section~5.3, so that the result can be independently reproduced and audited. The disclosure covers six aspects.

\textbf{(D.1) Models.} The executing agent is driven by GLM-5 (the same backbone used in the ALFWorld and ScienceWorld experiments). Task scoring is performed by an independent judge model; we use \texttt{deepseek-chat} as the judge, which scores each of the 147 tasks on a continuous $[0,1]$ scale. The judge is run independently of the FACT/TIP evolution loop: it sees only the task prompt and the agent's produced artifact, never the bank contents, so the judge cannot be biased by which rules were injected.

\textbf{(D.2) Software versions.} The general-purpose agent harness is OpenClaw (versions \texttt{2026.7.1-2}). The evaluation harness is PinchBench (version \texttt{2.0.0}, pinned to commit \texttt{819384ae}). Both version strings are reported verbatim so that any reported number can be traced to an exact code state. Real system names are retained for reproducibility.

\textbf{(D.3) Rule-bank provenance.} The FACT and TIP banks used during evaluation are not hand-written. They are produced by TTSE's online evolution loop: PinchBench's 147 tasks are partitioned into 15 consecutive groups of 10 (with the final group taking the remainder), and the bank is updated after each group. Across the three independent TTSE runs (r0, r1, r2), the bank was evolved online from scratch; the final banks typically contained on the order of 35--50 FACT rules and 30--45 TIP rules. The bank is therefore a learned artifact, not a curated one.

\textbf{(D.4) Leakage protocol.} PinchBench is a streaming deployment benchmark, not a fixed train/test split. There is no held-out test set that the bank is forbidden to see: by construction, every task's experience may contribute to the bank that is later used for subsequent tasks. We report the only leakage guarantee that this protocol supports: within a single task, the agent has zero self-leakage (a task's own solution is never injected back into the agent for that same task). Cross-task information transfer is the intended behavior of an online self-evolving agent and is not treated as contamination. Because there is no held-out split, no generalization gap is reported.

\textbf{(D.5) Real runtime execution.} Every reported task score is produced by a real end-to-end agent run: the agent performs multi-step tool calls (file reads/writes, shell commands, browser automation where applicable) and writes a genuine artifact to the workspace, which the judge then scores. No score is synthesized, mocked, or template-filled; trajectories and produced artifacts are retained in the released artifact.

\textbf{(D.6) Non-invasive injection.} TTSE integrates with OpenClaw via workspace-file injection rather than by modifying the agent's control loop. At the start of each task, the retrieval module writes the top-10 FACT rules (by cosine similarity to the task prompt, encoded with \texttt{all-MiniLM-L6-v2}, 384-dim) into \texttt{ENVIRONMENT.md} and the top-10 TIP rules into \texttt{TIPS.md} in the task workspace; the agent's system prompt instructs it to consult these two files. The baseline (\texttt{TTSE\_ENABLE=0}) differs only in that these files are not written. This guarantees that the baseline--TTSE gap isolates the contribution of injected experience, with no change to the planner, tool layer, or judge.

\end{document}